\documentclass[twoside]{article}

\usepackage[accepted]{aistats2020}

\usepackage[round]{natbib}

\usepackage[utf8]{inputenc} 
\usepackage[T1]{fontenc}    
\usepackage[hidelinks]{hyperref}       
\usepackage{url}            
\usepackage{booktabs}       
\usepackage{amsfonts}       
\usepackage{nicefrac}       
\usepackage{microtype}      

\usepackage{graphicx}
\usepackage{subfigure}
\usepackage{floatrow}
\usepackage{float}
\usepackage{wrapfig}
\usepackage[font=small]{caption}

\usepackage{appendix}

\usepackage{amsmath,amssymb}
\usepackage{amsfonts}
\usepackage{bm}

\usepackage{color}

\begin{document}

\runningtitle{POPCORN: Partially Observed Prediction Constrained Reinforcement Learning}

\setlength{\abovedisplayskip}{2pt plus 3pt}
\setlength{\belowdisplayskip}{2pt plus 3pt}

\twocolumn[

\aistatstitle{POPCORN: Partially Observed \\ Prediction Constrained Reinforcement Learning}

\aistatsauthor{Joseph Futoma \And Michael C. Hughes \And Finale Doshi-Velez}
\aistatsaddress{Harvard SEAS \And Tufts University, Dept. of Computer Science \And Harvard SEAS} 
 
]

\vspace{-0.25in}
\begin{abstract}
Many medical decision-making tasks can be framed as partially observed Markov decision processes (POMDPs).  
However, prevailing two-stage approaches that first learn a POMDP and then solve it often fail because the model that best fits the data may not be well suited for planning.  
We introduce a new optimization objective that (a) produces both high-performing policies and high-quality generative models, even when some observations are irrelevant for planning, and (b) does so in batch off-policy settings that are typical in healthcare, when only retrospective data is available. 
We demonstrate our approach on synthetic examples and a challenging medical decision-making problem.
\end{abstract}

\vspace{-0.2in}
\section{Introduction}
\vspace{-0.08in}

Reinforcement learning (RL) has the potential to assist sequential decision-making in healthcare, especially in settings lacking strong evidence-based guidelines.  
For example, in this work we will focus on the task of managing patients in an intensive care unit (ICU) with acute hypotension, a life-threatening emergency in which a patient's blood pressure drops dangerously low.  
In situations like this in critical care, it is often unclear which treatment will be most effective for a given patient, and in what amount, frequency, and duration~\citep{garcia2015effects}. 
RL might help answer these questions, but applying RL in healthcare is challenging, as highlighted by~\citet{gottesman2019GuidelinesReinforcementLearning}, and we are still far from integration into the clinic.  
Two key challenges for most clinical decision-making problems, including ours, are: 
\vspace{-0.1in}
\begin{enumerate} 
	\item Medical environments are \emph{partially observable}: a patient's current physiological state alone is insufficient to make good decisions, and we need other context about their history. 
	\item We must learn in a \emph{batch} setting, given only a single batch of retrospective (usually observational) data.
\end{enumerate}
\vspace{-0.1in}
In this work we focus on pushing the limits of \emph{model-based} RL, using \emph{discrete} hidden state representations. 
Generative models can of course help us learn to recommend good actions (our primary goal), but they also have many other important benefits. 
For instance, we can use them to predict future observations (a form of validation), they can learn in the presence of missing data (pervasive in clinical settings), and they are generally more sample-efficient than competing model-free approaches (important as many medical problems are data-limited). 
Building directly inspectable models via simple, discrete structures further enables easy inspection for clinical sensibility, a task much harder and sometimes impossible to accomplish from more complicated black-box models (e.g. deep learning).

We propose POPCORN\footnote{An implementation of POPCORN can be found online at \url{https://github.com/dtak/POPCORN-POMDP}, along with code to reproduce our experiments.}, or Partially Observed Prediction Constrained Reinforcement learning, a new optimization objective for the well-known partially observable Markov decision process (POMDP)~\citep{kaelbling1998}.  
POMDPs have been traditionally trained in a two-stage process, where the first stage learns a generative model by maximizing the likelihood of observations and is not tied to the decision-making task. 
However, this approach can fail to find good policies when the model is (inevitably) misspecified; in particular, a maximum likelihood model may spend capacity modeling irrelevant information rather than signal important for the task at hand.  
We demonstrate this effect and show how POPCORN, which constrains maximum likelihood training of the POMDP model so that the value of the model's induced policy achieves satisfactory performance, can overcome these issues.  

\vspace{-.1in}
\section{Related Work}
\label{sec:related_work}
\vspace{-.1in}

\textbf{RL in Healthcare.}
Healthcare applications of RL have proliferated in recent years, in diverse clinical areas such as schizophrenia~\citep{shortreedInformingSequentialClinical2011}, sepsis~\citep{komorowski2018artificial,raghuContinuousStateSpaceModels2017a}, mechanical ventilation~\citep{prasadReinforcementLearningApproach2017}, HIV~\citep{ernst2006clinical}, and dialysis~\citep{martin-guerreroReinforcementLearningApproach2009}. However, most works use model-free approaches, ostensibly because learning accurate models from noisy biological data is challenging.  All of these works further assume full-observability, which is often not accurate.

A few prior works in this applied space have explicitly modeled partial observability. POMDPs have been applied to heart disease management~\citep{hauskrechtPlanningTreatmentIschemic2000}, sepsis treatment in off-policy or simulated settings~\citep{liActorSearchTree2018,oberstCounterfactualOffPolicyEvaluation2019,peng2018improving}, and HIV management~\citep{parbhoo2017combining}.  All of these approaches take a two-stage approach to learning.  In contrast, our approach is \emph{decision-aware} throughout the optimization process.

\textbf{Imperfect Models in RL}
Model-based RL is a long-standing area of research, and work as early as \citet{abbeel2006} looked at learning misspecified models that are still useful for RL.
More broadly, ``end-to-end'' optimization methods directly incorporate a downstream decision-making task during model training, and are growing in popularity across machine learning, from graphical models~\citep{lacoste-julienApproximateInferenceLosscalibrated2011} to submodular optimization~\citep{wilderMeldingDataDecisionsPipeline2019}.  
Within RL, recent decision-aware optimization efforts have explored partially-observed problems in both model-free~\citep{QMDP-net} and model-based settings~\citep{Igl2018}.

These RL efforts differ from ours in two key respects. First, they exclusively focus on on-policy settings for simulated environments such as Atari. Second, they rely heavily on black-box neural networks for feature extraction, which are not generally sample-efficient or easily interpreted.  In many cases (e.g.~\citet{QMDP-net}), the model is treated as an abstraction and there is no way to set the importance of the model's ability to accurately generate trajectories.  Perhaps closest in spirit to our approach is theoretical work on value-aware model learning in RL~\citep{farahmand2018iterative}.

\vspace{-0.1in}
\section{Background}
\label{sec:background}
\vspace{-0.1in}

\textbf{POMDP Model.}
We consider a POMDP with $K$ discrete latent states (e.g. physiological conditions of patients), $A$ discrete actions (e.g. possible treatments), $D$-dimensional observations (e.g. clinical measurements), and deterministic rewards (e.g. how ``good'' or ``bad'' the treatments were).
The entire generative model for states $s_t \in \{1, 2, \ldots K \}$ and observations $o_t \in \mathbb{R}^D$ across timesteps $t \in \{0, 1, \ldots T\}$ is given by:
\begin{align}
\label{eq:iohmm}
	p(s_0=k)
        &\triangleq \tau_{0k}, \\ \notag
    p(s_{t+1}=k | s_t=j, a_t=a)
        &\triangleq \tau_{ajk}, \\ \notag
    p(o_{t+1,d} | s_{t+1}\text{=}k, a_t\text{=}a)
    &\triangleq \mathcal{N}(\mu_{akd}, \sigma_{akd}^2 ). \notag
\end{align}
We define the model parameters as $\theta \equiv \{ \tau, \mu, \sigma, R\}$.
$\tau$ describes the transition probability of moving to the next (unobserved) state $s_{t+1}$, given current state $s_t$ and action $a_t$. 
We model each observation $o_t$ as Gaussian, with emission parameters $\mu$ and $\sigma^2$ denoting the mean and variance when in state $s_t$ after taking action $a_{t-1}$.
Although any (tractable) distribution is possible, we choose to use independent Gaussians across the $D$ dimensions for simplicity.
Completing the POMDP specification is the deterministic reward function $R(s,a)$, specifying the reward from taking action $a$ in state $s$.

A dataset consists of $N$ trajectories (e.g. decisions made about a patient's care, along with clinical observations). We index each trajectory by $n \in \{1, \ldots, N\}$, with the length of trajectory $n$ given by $T_n \leq T$.

Given a POMDP with parameters $\theta$, we can compute the \emph{belief} $b_t \in \Delta^K$,
a vector in the simplex $\Delta^K \triangleq \{ q \in \mathbb{R}^K | \sum_{k=1}^K q_k = 1, q_k \geq 0\}$.
The belief defines the posterior over state $s_t$ given all past actions and observations: $b_{tk} \triangleq p(s_t=k | o_{0:t}, a_{0:t-1})$, is a sufficient statistic for the entire history, and can be computed efficiently via forward filtering~\citep{rabiner1989hmm}. We can \emph{solve} the POMDP using a planning algorithm to learn a policy $\pi_\theta: \Delta^K \to \Delta^A$, mapping any  belief to a distribution over actions (or a single action for deterministic policies). The goal is to find a policy with high value (the expected sum of discounted rewards): $V^\pi = \sum_{t=0}^T \gamma^t \mathbb{E}[ r_t ]$, with $\gamma \in (0,1)$ the discount.

\textbf{Learning Parameters: Input-Output HMM.}
The model in Eq.~\eqref{eq:iohmm} is an input-output hidden Markov model (IO-HMM)~\citep{bengio1995ioHMM}, where actions are inputs and observations are outputs.  
The model parameters $\{\tau, \mu, \sigma\}$ that maximize the marginal likelihood of observed trajectories can be efficiently computed using the EM algorithm for HMMs~\citep{rabiner1989hmm,chrisman1992baumwelchForPOMDPs}. 
For Bayesian approaches, efficient algorithms for sampling from the posterior over POMDP models also exist \citep{doshivelezThesis2012}. The deterministic reward function $R$ is estimated separately by minimizing squared errors with the observed rewards (see Appendix \ref{app:learning-rewards} for additional details).

\textbf{Solving for the Policy.}
The value function of a discrete-state POMDP can be modeled arbitrarily closely as the upper envelope of a finite set of linear functions of the belief \citep{sondik1978pomdp}. However, exact value iteration is intractable even for very small POMDPs.  In this work, we use point based value iteration (PBVI) \citep{Pineau2003}, an approximate algorithm that is significantly more efficient (see \citet{shani2013survey} for a survey of related algorithms and extensions). Rather than perform Bellman backups over all valid beliefs $b \in \Delta^K$, PBVI only performs backups at a specific (small) set of beliefs. For the modest state spaces in our applications ($K << 100$), PBVI is an efficient solver. However, we require two key innovations beyond standard PBVI. First, we adapt ideas from \citet{Hoey2005} to handle continuous observations. Second, we relax the algorithm so that each step is differentiable. See Appendix \ref{app:pbvi} for details.

\textbf{Off-Policy Value Estimation.}
Let $\pi_{\theta}$ be the (near-optimal) policy obtained from PBVI for the given model parameters $\theta = \{\tau, \mu, \sigma, R\}$. The fact that $\pi_{\theta}$ is optimal for a specific model $\theta$ does not mean it is optimal in practice (e.g. in the clinic), because our generative model is almost certainly misspecified. If we have access to an environment simulator, we can evaluate $\pi_\theta$ via standard Monte Carlo rollouts. However, in the \emph{batch} setting, we lack the ability to interact with the environment and must turn to \emph{off policy evaluation} (OPE) to \emph{estimate} a policy's value.  

Let $\pi_{\text{beh}}$ denote the behavior policy under which the observed data were collected (e.g. clinician treatment tendencies).\footnote{In our experiments with simulated environments, we assume $\pi_{beh}$ is given. In the real data setting, we estimate the behavior policy via the k-nearest neighbors approach of \citet{raghu2018behaviour}. See Appendix \ref{app:mimic:behavior} for details.} Let $\mathcal{D}$ denote a set of $N$ trajectories collected under $\pi_{\text{beh}}$. In this work, the specific OPE technique we  use is consistent weighted per-decision importance sampling (CWPDIS, \citep{thomas2015phdthesis}) which estimates the value of a  policy $\pi_{\theta}$ as:
\begin{align}
\label{eq:cwpdis}
   V^{\text{CWPDIS}}(\pi_{\theta}) & \triangleq
   \sum_{t=1}^T \gamma^{t}
    \frac
        {
        \sum_{n \in \mathcal{D}}
            r_{nt}
            \rho_{nt}( \pi_\theta )
        }
        {
        \sum_{n \in \mathcal{D}}
            \rho_{nt}( \pi_\theta )
        }, \\ \notag
    \rho_{nt}(\pi_\theta) &\triangleq \prod_{s=0}^t
        \frac
        {\pi_{\theta}(a_{ns}|o_{n,0:s}, a_{n,0:s-1})}
        {\pi_{\text{beh}}(a_{ns}|o_{n,0:s},a_{n,0:s-1})}.
\end{align}
In general, importance sampling (IS) estimators such as CWPDIS have lower bias than other approaches but suffer from higher variance. Another class of OPE methods learn a separate model to simulate trajectories in order to estimate policy values (e.g. \citet{chow2015robust}), but may suffer from unacceptably high bias in real-world, noisy settings. 

\vspace{-0.1in}
\section{Prediction-Constrained POMDPs}
\label{sec:POPCORN}
\vspace{-0.1in}

We now introduce POPCORN, our proposed \emph{prediction-constrained} optimization framework for learning POMDPs. We seek to learn parameters $\theta$ that will both assign high likelihood to the observed data while also yielding a policy $\pi_{\theta}$ with high (estimated) value.
As noted in Sec.~\ref{sec:related_work}, previous approaches for learning POMDPs generally fall into two categories.  Two-stage methods (e.g.~\citet{chrisman1992baumwelchForPOMDPs}) that first learn a model and then solve it often fail to find good policies under severe model misspecification.  End-to-end methods (e.g. ~\citet{QMDP-net}) that focus only on the ``discriminative'' task of policy learning typically fail to produce accurate generative models of the environment. They also lack the ability to handle missing observations, which is especially problematic in medical contexts where missing data is pervasive.

Our approach offers a \emph{balance} between these purely maximum likelihood-driven (generative) and purely reward-driven (discriminative) extremes. We retain the strengths of the generative approach---the ability to plan under missing observations, simulate accurate dynamics, and inspect model parameters to inspire scientific hypotheses---while benefiting from model parameters that are directly informed by the decision task as in end-to-end frameworks.

\vspace{-0.1in}
\subsection{POPCORN Objective}
\vspace{-0.1in}

Our proposed framework seeks a POMDP $\theta$ that maximizes the log marginal likelihood $\mathcal{L}_{\text{gen}}$ of the observed data $\mathcal{D}$, while enforcing that the solved policy's (estimated) value $V(\pi_\theta)$ is high enough to be useful.
Formally, we seek a $\theta$ that maximizes the constrained optimization problem:
\begin{align}
\max_{\theta} \mathcal{L}_{\text{gen}}( \theta ), \qquad \text{subject to:}~ V(\pi_\theta) \geq \epsilon,
\label{eq:pc-pomdp-objective-constrained}
\end{align}
with the functions $\mathcal{L}_{\text{gen}}$ and $V$ defined below. 
The tolerance $\epsilon$ defines a minimum acceptable policy value (e.g. as determined by a domain expert).

Setting practical optimization considerations aside, we would prefer the constrained formulation of Eq.~\eqref{eq:pc-pomdp-objective-constrained} as it best expresses our model-fitting goals: as good a generative model as possible, but we will not accept poor decision-making.
This objective is similar to the \emph{prediction-constrained} objective used by~\citet{hughes2018pc} for optimizing supervised topic models; here we apply similar ideas to batch, off-policy RL settings.  

In practice, solving constrained problems is challenging, so we transform to an equivalent unconstrained objective using a Lagrange multiplier $\lambda > 0$:
\begin{align}
\max_{\theta} \quad \mathcal{L}_{\text{gen}}( \theta ) + \lambda V(\pi_\theta).
\label{eq:pc-pomdp-objective-unconstrained}
\end{align}
Setting $\lambda = 0$ recovers classic two-stage training, while the limit $\lambda \rightarrow \infty$ approximates end-to-end approaches. In our experiments, we compare against both of these baseline approaches, referring to the $\lambda=0$ case as ``2-stage'', and the  $\lambda \rightarrow \infty$ case as ``Value-only'' (for this case, in practice we optimize $V(\pi_\theta)$ and ignore $\mathcal{L}_{\text{gen}}$).

\textbf{Computing the Generative Term.}
We define $\mathcal{L}_{\text{gen}}(\theta)$ as the log marginal likelihood of observations, given the actions in $\mathcal{D}$ and parameters $\theta$:
\begin{equation} 
\mathcal{L}_{\text{gen}}(\theta)=
\textstyle \sum_{n \in \mathcal{D}}
\log p(o_{n,0:T_n} | a_{n,0:T_n-1}, \theta).
\end{equation} 
This IO-HMM likelihood marginalizes over uncertainty about the hidden states, can be computed efficiently via dynamic programming~\citep{rabiner1989hmm}, and is also amenable to automatic differentiation w.r.t. $\theta$.

\textbf{Computing the Value Term.} 
Computation of $V(\pi_{\theta})$ entails two distinct parts: solving for the policy $\pi_{\theta}$ given $\theta$, and then estimating the value of this policy using OPE and $\mathcal{D}$.  We require both to be differentiable to permit gradient-based optimization. 
To solve for the policy, we apply a differentiable relaxation of PBVI (see Appendix \ref{app:pbvi:softmax-relax} for full details). Although standard PBVI returns a deterministic policy, we relax this as well and learn stochastic policies as they are generally easier to evaluate with OPE. We emphasize that our framework is general and other solvers are possible as long as they can be made differentiable. 
To compute the estimated policy value, we use the CWPDIS estimator in Eq.~\eqref{eq:cwpdis}.  As it is a \emph{differentiable} function of $\theta$, our unconstrained objective in Eq.~\eqref{eq:pc-pomdp-objective-unconstrained} can be optimized via first-order gradient ascent.

\vspace{-0.1in}
\subsection{Optimizing the Objective}
\vspace{-0.1in}

We optimize using gradients computed from the full dataset (we do not use subsample to avoid extra variance). We optimize with Rprop \citep{rprop} with default settings. Our objective is challenging due to non-convexity, as even the generative term alone admits many local optima. To improve solution quality in all experiments and for all methods, prior to final evaluation we take the best of 25 random restarts as measured by training objective value.

\textbf{Stabilizing the Off-Policy Estimate.}
Although our OPE estimate (using CWPDIS) was reliable in simulated environments, on our real dataset it had unusably high variance, as is common with IS estimators. We address this in two ways. 

First, we add an extra term in the objective encouraging larger effective sample size (ESS) and hence lower variance, following \citet{metelliPolicyOptimizationImportance2018}. Our final objective includes an ESS penalty with weight $\lambda_{\text{ESS}} > 0$:
\begin{align}
    \max_{\theta} \quad \mathcal{L}_{\text{gen}}(\theta) + \lambda \cdot \left[ V(\pi_\theta) .- \frac{\lambda_\text{ESS}}{\sqrt{\text{ESS}(\theta)} } \right].
\end{align}
As the CWPDIS estimator in Eq.~\ref{eq:cwpdis} is the weighted sum of a sequence of $T$ IS estimators (the average discounted reward at each $t$), we sum all these stepwise $\text{ESS}_t$ values to yield the final $\text{ESS}(\theta)$ term.
$\text{ESS}_t$ is approximated given IS weights $\{\rho_{nt}\}_{n=1}^N$ as $\frac{(\sum_n \rho_{nt})^2}{\sum_n \rho_{nt}^2}$~\citep{kongNoteImportanceSampling1992}.

Second, we restrict the support of $\pi_\theta$ and then renormalize to only allow actions where there was at least $\delta$ probability under $\pi_{\text{beh}}$.
This forces strong overlap between the support of $\pi_{\text{beh}}$ and $\pi_\theta$ and often substantially reduces the variance of the final OPE estimate. 
This also provides a soft notion of ``safety'', as now rare or unknown actions are prohibited. 

\textbf{Hyperparameters.}
The key hyperparameter for POPCORN is the scalar tradeoff $\lambda > 0$.  We try a range of 5 $\lambda$'s per environment spaced evenly on a log scale. 
We also rescale $\mathcal{L}_{\text{gen}}(\theta)$ by the total number of observed scalars ($D (\sum_n T_n)$ if there is no missing data), so that the magnitude of $\lambda$ has roughly consistent impact across datasets. We also try 5 ESS penalty weights $\lambda_{\text{ESS}}$ evenly spaced on a log scale, but this term was only necessary for the real data experiments.

\vspace{-0.1in}
\section{Simulated Environments}
\label{sec:experimental_results} 
\vspace{-0.1in}

We first evaluate POPCORN on three simulated environments to validate its utility across a range of possible model misspecification scenarios. We later evaluate on a more difficult medical simulator. For all experiments in this section, everything is conducted in the batch, off-policy setting. The simulator is only used to produce the initial data set and to evaluate the final policy after training concludes.  We separate each experiment into a description of procedure and highlights of the results.

Recall that our goal is to learn simple---and therefore interpretable---models that perform robustly in misspecified settings. As such, we compare against an approach that does not attempt to model the dynamics (``value term only''), an approach that first learns the model and then plans (``2-stage''), and a known optimal solution (when available).  In all cases, we are interested in how these methods trade off between explaining the data well (log marginal likelihood of data) and making good decisions (policy value).

\vspace{-0.1in}
\subsection{Synthetic Domains with Misspecification}
\vspace{-0.1in}

\begin{figure*}[!t]
\begin{tabular}{c c c}
\includegraphics[height=0.18\textwidth]{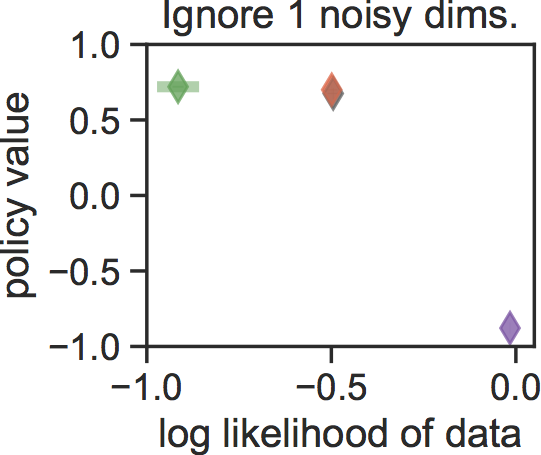}
&
\includegraphics[height=0.18\textwidth]{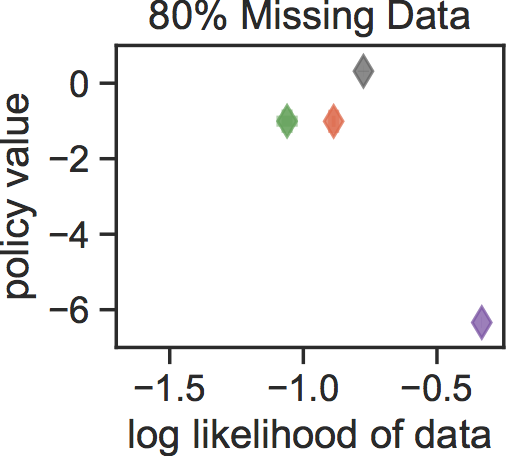}
&
\includegraphics[height=0.18\textwidth]{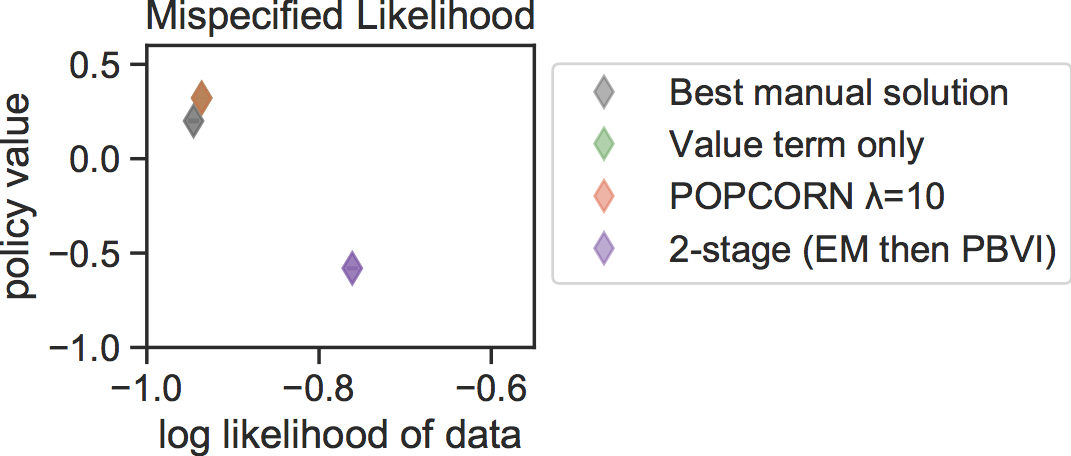}
\end{tabular}
    \caption{ 
    Solutions from all competitor methods in the 2D fitness landscape (policy value on y-axis; log marginal likelihood on x-axis). 
    An ideal method would score in the top right corner of each plot.
    \emph{Left}: Results from Tiger with Irrelevant Noise Dimensions.
    \emph{Middle}: Results from Tiger with Missing Data.
    \emph{Right}: Results from Tiger with Wrong Likelihood.
    POPCORN is robust to all three types of model misspecification tested, and consistently learns better policies than 2-stage and better models than value-only.
    }
    \label{fig:Tiger_results}
\end{figure*}

We demonstrate how POPCORN overcomes various kinds of model misspecification in the context of the classic POMDP tiger problem \citep{kaelbling1998}.  The tiger problem consists of a room with 2 doors: one door is safe, and the other door has a tiger behind it. The agent has 3 possible actions: either open one of the doors, thereby ending the episode, or listen for noisy evidence of which door is safe to open. Revealing a tiger gives $-5$ reward, the safe door yields $+1$ reward, and listening incurs $-0.1$ reward. The goal is to maximize rewards over many repeated trials, with the ``safe'' door's location randomly chosen each time.  

We set $\gamma = 0.9$ to encourage the agent to act quickly.
We collect training data from a random policy that first listens for 5 time steps, and then randomly either opens a door or listens again.
We train in the \emph{batch setting} given a single collection of 1000 trajectories of length 5-15.
After optimization, we evaluate each policy via an additional 1000 Monte Carlo rollouts.

\textbf{Tiger with Irrelevant Noise: Finding dimensions that signal reward.}
In this setting, whenever the agent listens for the tiger, it receives an observation $o_t$ with $D = 2$ dimensions. The first dimension provides a noisy signal as to the location of the safe door.
We set this ``signal'' dimension $o_{t1} \sim \mathcal{N}(i_{\text{safe}}, 0.3^2)$, where the mean is the safe door's index $i_{\text{safe}} \in \{0,1\}$.
The second dimension is irrelevant to the safe door's location, and we set $o_{t2} \sim \mathcal{N}(j, 0.1^2)$, with $j \sim \mbox{Unif}(\{0,1\})$ in each trial. Thus, $K=4$ total states would be needed to explain perfectly both the relevant and irrelevant signals for all possible values of $(i_{\text{safe}}, j)$.

We measure performance allowing only $K=2$ states to assess how each method spends its limited capacity across the generative and reward-seeking goals. 
We expect the 2-stage baseline will identify the irrelevant states indexed by $j$, as they have lower standard deviation (0.1 vs. 0.3 for the signal dimension) and thus are more important to maximize likelihood. In contrast, we expect POPCORN will focus on the relevant signal dimension and recover high-value policies.

\textbf{Tiger with Missing Data: Finding relevant dimensions when some data is missing.}
This next scenario extends the previous setting in which the listen action produces $D=2$ observations, where the first signals the safe door's location and the second is irrelevant.  However, now the dimension with the relevant signal is often missing.  Specifically, $o_{t1} \sim \mathcal{N}(i_{\text{safe}}, 0.3^2)$ and $o_{t2} \sim \mathcal{N}(j, 0.3^2)$, but we select 80\% of signal observations $o_{t1}$ to be \emph{missing} uniformly at random.  This (coarsely) simulates clinical settings where some measurements may be infrequent but important (e.g. relevant lab tests), while others are common but not directly useful (e.g. routine vitals).

The expected outcome with $K=2$ states is that a 2-stage approach driven by maximizing likelihood would prefer to model the always-present irrelevant dimension. In contrast, POPCORN should learn to favor the signal dimension even though it is rarely available and contributes less overall to the likelihood.  This ability to gracefully handle missing dimensions is a natural property of generative models and would not be easily done with a model-free approach.

\textbf{Tiger with Wrong Likelihood: Overcoming a misspecified density model.}
Finally, we consider a situation in which our generative model's \emph{density family} cannot match the true observation distribution.  This time, the listen action produces a $D=1$ dimensional observation $o_t$.  The true distribution of this observation signal is a \emph{truncation} of a mixture of two Gaussians, denoted $\text{GMM}(o) = 0.5 \mathcal{N}(o | 0, 0.1^2) + 0.5\mathcal{N}(o | 1, 1.0^2)$. If the first door is safe, listening results in strictly negative observations: $p(o) \propto \delta(o < 0) \text{GMM}(o)$. If the second door is safe, listening results in strictly positive observations: $p(o) \propto \delta(o > 0) \text{GMM}(o)$.

While the the true observation densities are \emph{not} Gaussian, we will fit POMDP models with Gaussian likelihoods and $K=2$ states.  We expect POPCORN to still deliver high-value policies, even though the likelihood will be suboptimal. 
See Appendix \ref{app:tiger} for more details on the overall setup of all three tiger environments as well as additional results.

\vspace{-0.1in}
\subsection{Conclusions from Synthetic Domains}
\vspace{-0.1in}

Across all variants of the Tiger problem, we observe many common conclusions from Fig.~\ref{fig:Tiger_results}.  Together, these results demonstrate how POPCORN is robust to many different kinds of model misspecification.  

\textbf{POPCORN learns consistently better policies than 2-stage.}
Across all 3 panels of Fig.~\ref{fig:Tiger_results}, POPCORN (red) delivers higher value $V(\pi_{\theta})$ (y-axis) than the 2-stage baseline (purple).

\textbf{Value-only learns poor generative models.}
In 2 of 3 panels, the value-only baseline (green) has noticeably worse likelihood $\mathcal{L}_{\text{gen}}(\theta)$ (x-axis) than POPCORN. The far right panels shows indistinguishable performance. Notably, optimizing this objective is significantly less stable than the full POPCORN objective. This aligns with findings from \citet{levine2013guided}, who also observed that policy learning via direct optimization of IS estimates of policy value is challenging.

\textbf{POPCORN solutions are consistent with manually-designed solutions.}
In all 3 panels, POPCORN (red) is the closest method to the ideal manually-designed solution (gray).

\vspace{-0.1in}
\subsection{Sepsis Simulator: Medically-motivated environment with known ground truth.}
\vspace{-0.1in}

We now move from simple toy problems---each designed to demonstrate a particular robustness of our method---to a more challenging simulated domain.  In real-world medical decision-making tasks, it is impossible to evaluate the value of a learned policy using data collected under that policy's decisions. However, in a simulated setting, we can evaluate any given policy to assess its true value.
We emphasize $\theta$ is still learned in the batch setting, as only after optimization do we use the simulator to allow for accurate evaluation of policy values.
	
We use the simulator from \citet{oberstCounterfactualOffPolicyEvaluation2019}, which is a coarse physiological model for sepsis with $D=5$ discrete observations: 4 ordinal-valued vitals (e.g. ``low''/``normal''/``high''), and a binary indicator for diabetes. The true simulator is governed by an underlying known Markov decision process (MDP), which has 1440 possible discrete states. There are 8 actions (3 different binary actions), and trajectories are at most 20 timesteps. Rewards are sparse, with 0 reward at intermediate time steps and $-1$ or $+1$ at termination.

To make this simulator similar to our other environments with continuous-valued observations, we add independent Gaussian noise with standard deviation $0.3$ to each observation. This measurement error also makes the environment partially observable so that modeling it as a POMDP is reasonable. Although \citet{oberstCounterfactualOffPolicyEvaluation2019} used structural causal models to simulate counterfactual trajectories and explicitly address causal questions, our POMDP construction implicitly assumes no hidden confounding. Our work skirts causality, as we view POMDPs solely as a convenient way to summarize trajectory histories. Our use of this simulator hence differs substantially from its original use, where it was used to create strong (known) hidden confounding in order to illustrate failure modes of OPE\footnote{We refer interested readers to recent related work in \citet{tennenholtz2020} that specifically proposes a new technique for performing OPE with POMDPs where there is hidden confounding. This differs from \citet{oberstCounterfactualOffPolicyEvaluation2019} who focus solely on a new approach to generating counterfactual trajectories under a proposed policy.}. 

The true discrete-state MDP is easily solved via exact value iteration. We generate $2500$ trajectories under an $\epsilon$-greedy behavior policy, with $\epsilon=0.14$ so each non-optimal action has a $.02$ probability of being taken. Given observed trajectories, we learn POMDPs assuming $K=5$ (we obtained similar qualitative results for other $K$), and evaluate policies via an additional $2500$ Monte Carlo rollouts. See Appendix \ref{app:sepsis} for full details.

\begin{figure}[t!]
 \includegraphics[width=\textwidth]{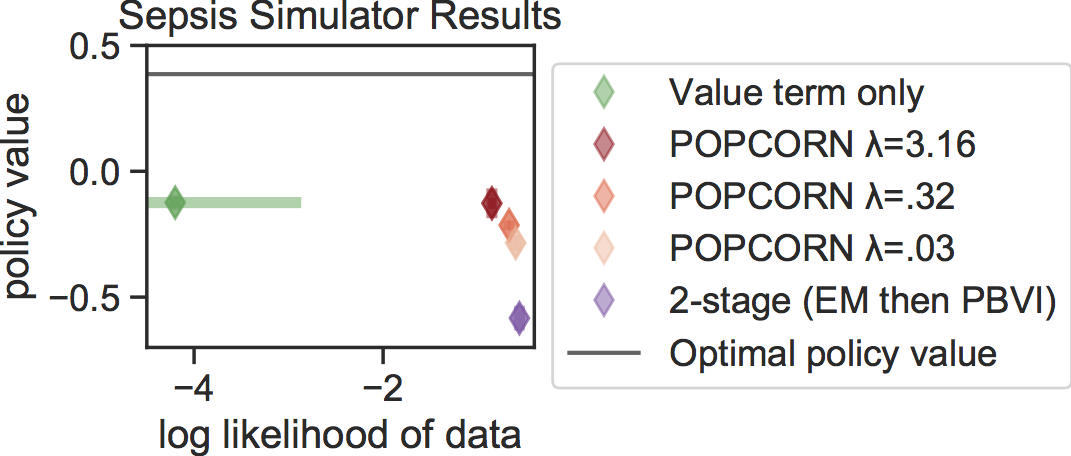}
\caption{
Quantitative results from the sepsis simulator. An ideal method would score in the top right corner. No methods recover a fully optimal policy (grey line), but POPCORN consistently learns better policies than 2-stage and better models than value-only. A policy which takes actions uniformly at random has a value of $-0.72$, so the 2-stage policy barely outperforms this.
}
\label{fig:sepsis}
\end{figure}

\textbf{Results and Conclusions.}
Figure~\ref{fig:sepsis} shows POPCORN again learns higher-value policies than 2-stage.
Additionally, while the value-term-only baseline learns a policy on par with POPCORN, its likelihood is substantially lower. While all policies are far from optimal, this unsurprising given the small state space, modest observation noise, and high $\epsilon$ for the behavior policy.

\vspace{-0.1in}
\section{Real Data Application: Hypotension}
\label{sec:hypotension_application}
\vspace{-0.1in}

\begin{figure}[!t]
\centering
\setlength{\tabcolsep}{0.01cm}
	\includegraphics[width=\textwidth]{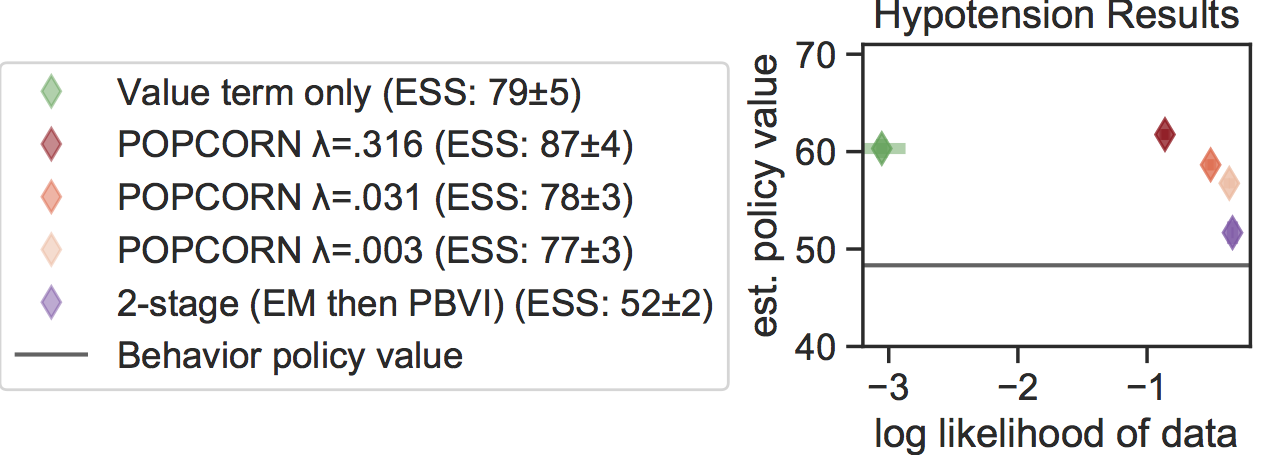}
\caption{
Quantitative results from the hypotension data, showing the trade-offs between policy value and likelihood (ESS is in legend). POPCORN again learns much better models than value-only and better policies than 2-stage.
}
\label{fig:hypo-tradeoff}
\end{figure}

To showcase the utility of our method on a real-world medical decision making task, we apply POPCORN to the challenging problem of managing acutely hypotensive patients in the ICU. Although hypotension is associated with high morbidity and mortality \citep{jones2006emergency}, management of these patients is difficult and treatment strategies are not standardized, in large part because there are many underlying potential causes of hypotension. Previosuly, \citet{girkar2018predicting} attempted to predict the efficacy of fluid therapy for hypotensive patients with only mixed success.  We apply POPCORN to this problem and first study the same trade-offs between generative and reward-seeking performance as in Sec.~\ref{sec:experimental_results}. We further perform an in-depth evaluation of the learned policy and our confidence in it (via effective sample sizes and qualitative checks).

\textbf{Cohort.} We use 10,142 ICU stays from MIMIC-III \citep{johnson2016mimic}, filtering to adult patients with at least 3 abnormally low mean arterial pressure (MAP) values in the first 72 hours of ICU admission. Our observations consist of 9 vitals and laboratory measurements: MAP, heart rate, urine output, lactate, Glasgow coma score, serum creatinine, FiO$_2$, total bilirubin, and platelets count. We discretized time into 1-hour windows, and setup the RL task to begin 1 hour after ICU admission to ensure a sufficient amount of data exists before starting a policy. Trajectories end either at ICU discharge or at 72 hours into the ICU admission, so there are at most 71 actions taken. This formulation was made in consultation with a critical care physician, who advised most acute cases of hypotension would present early during an ICU admission. We expressly do \emph{not} impute missing observations: only observed measurements contribute to the overall likelihood. 

\textbf{Setup.} Our action space consists of the two main treatments for acute hypotension: fluid bolus therapy and vasopressors, both of which are designed to quickly raise blood pressure and increase perfusion to the organs. We discretize fluids into 4 actions (none, low, medium, high), and discretize vasopressors into 5 actions (none, low, medium, high, very high) for a total of 20 discrete actions. 
To assign rewards to individual time steps, we use a piecewise-linear function created in consultation with a critical care physician. A MAP of 65mmHg is a common target~\citep{asfar2014high}, so if an action is taken and the next MAP is 65 or higher, the next reward is +1, the highest possible value. Otherwise, rewards decrease as MAP values drop, with MAP $\leq 30$ delivering a reward of 0, the smallest possible value. 
Further details on the action space discretization, a plot of the reward function, and other preprocesssing can be found in Appendix \ref{app:mimic}.

We split the dataset into 5 distinct test sets for cross-validation, and throughout present results on the test sets, with standard errors across folds where appropriate. We set $\lambda_{\text{ESS}}=4$ and set $\delta=0.03$, which prohibits all actions assigned less than $3\%$ probability by our estimated behavior policy.  Lastly, we study several possible values for the Lagrange multiplier, $\lambda \in \{10^{-2.5}, 10^{-1.5}, 10^{-0.5}\}$.

\begin{figure*}[!t]
\centering
\setlength{\tabcolsep}{0.01cm}
\begin{tabular}{c c c}
  \includegraphics[height=.14\textwidth]{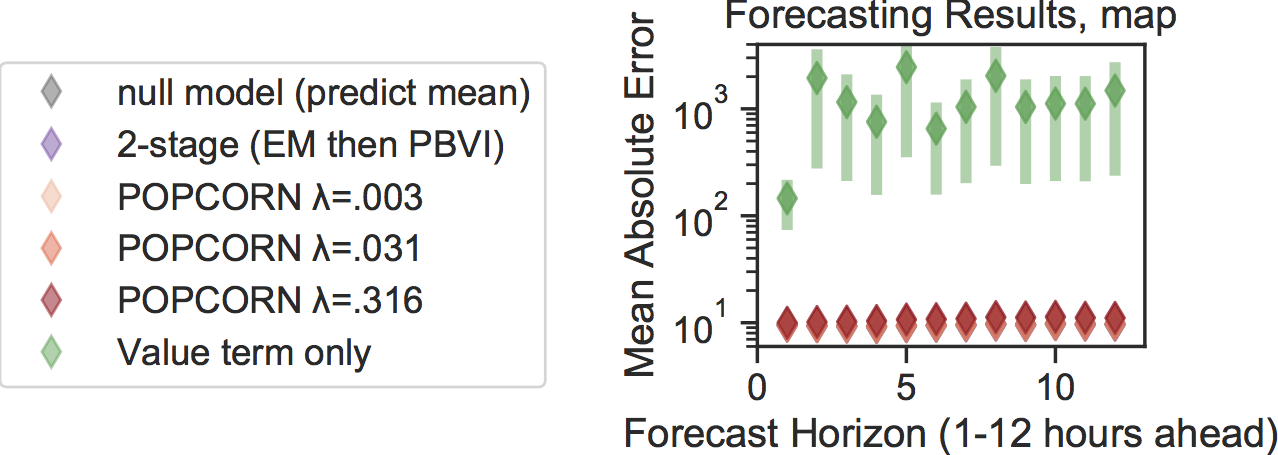}
&
  \includegraphics[height=.14\textwidth]{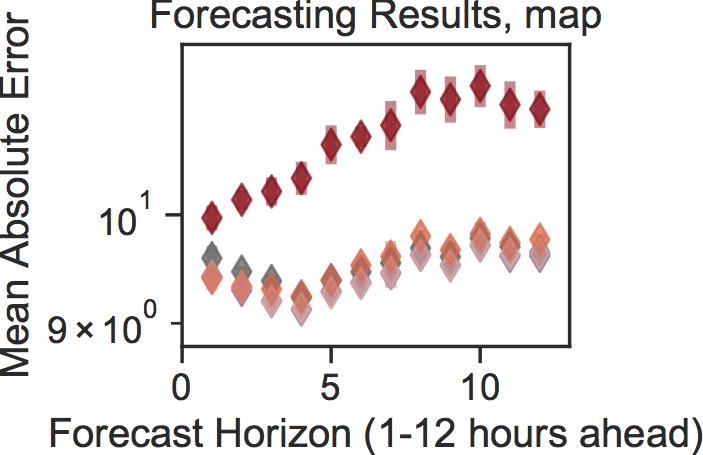}
&
  \includegraphics[height=.14\textwidth]{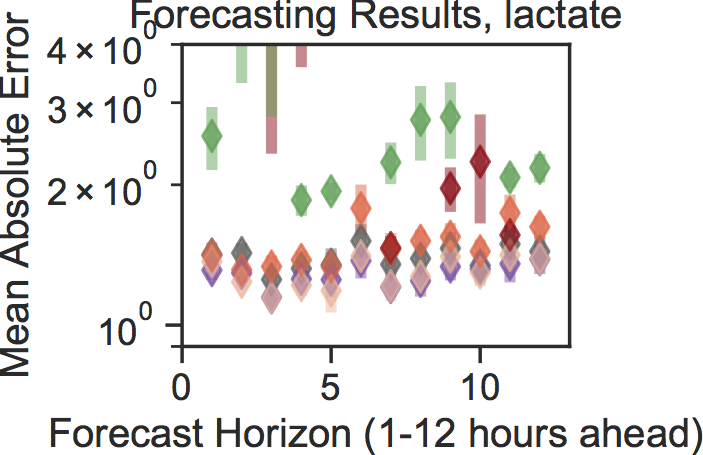}
\\
  \includegraphics[height=.14\textwidth]{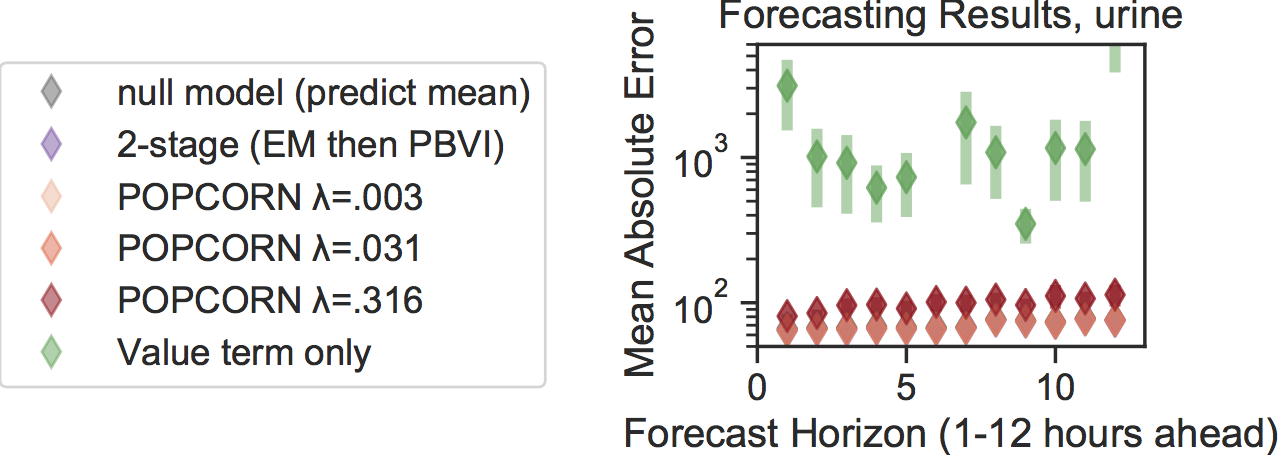}
&
  \includegraphics[height=.14\textwidth]{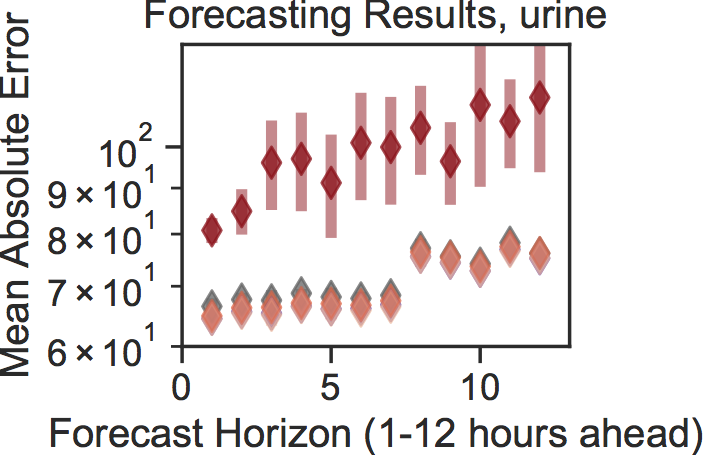}
&
  \includegraphics[height=.14\textwidth]{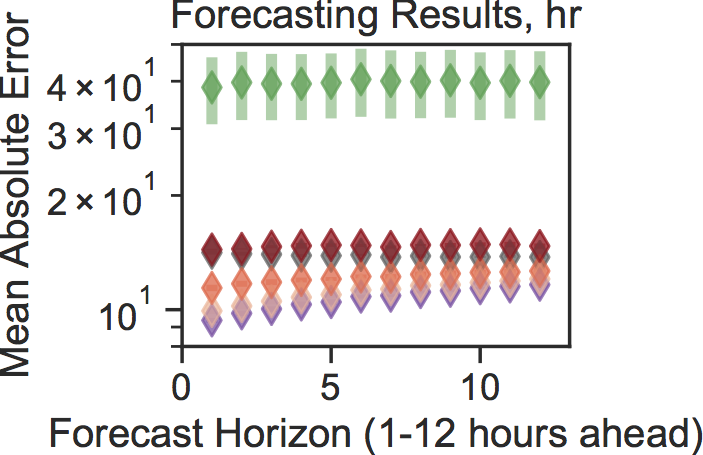}
\end{tabular}
\caption{Forecasting results. \emph{Top to bottom, left to right}: MAP (scale zoomed out); MAP (value-only out of pane); lactate; urine output (scale zoomed out); urine output (value-only out of pane); heart rate. 2-stage performs the best throughout, but for smaller values of $\lambda$ POPCORN is often not much worse. Value-only constantly makes wildly inaccurate predictions, as its forecast errors are often several orders of magnitude worse (see MAP and urine results in first column).}
\label{fig:hypo-forecast}
\end{figure*}

\begin{figure*}[!t]
\centering
\setlength{\tabcolsep}{0.01cm}
\begin{tabular}{c c c}
  \includegraphics[height=0.185\textwidth]{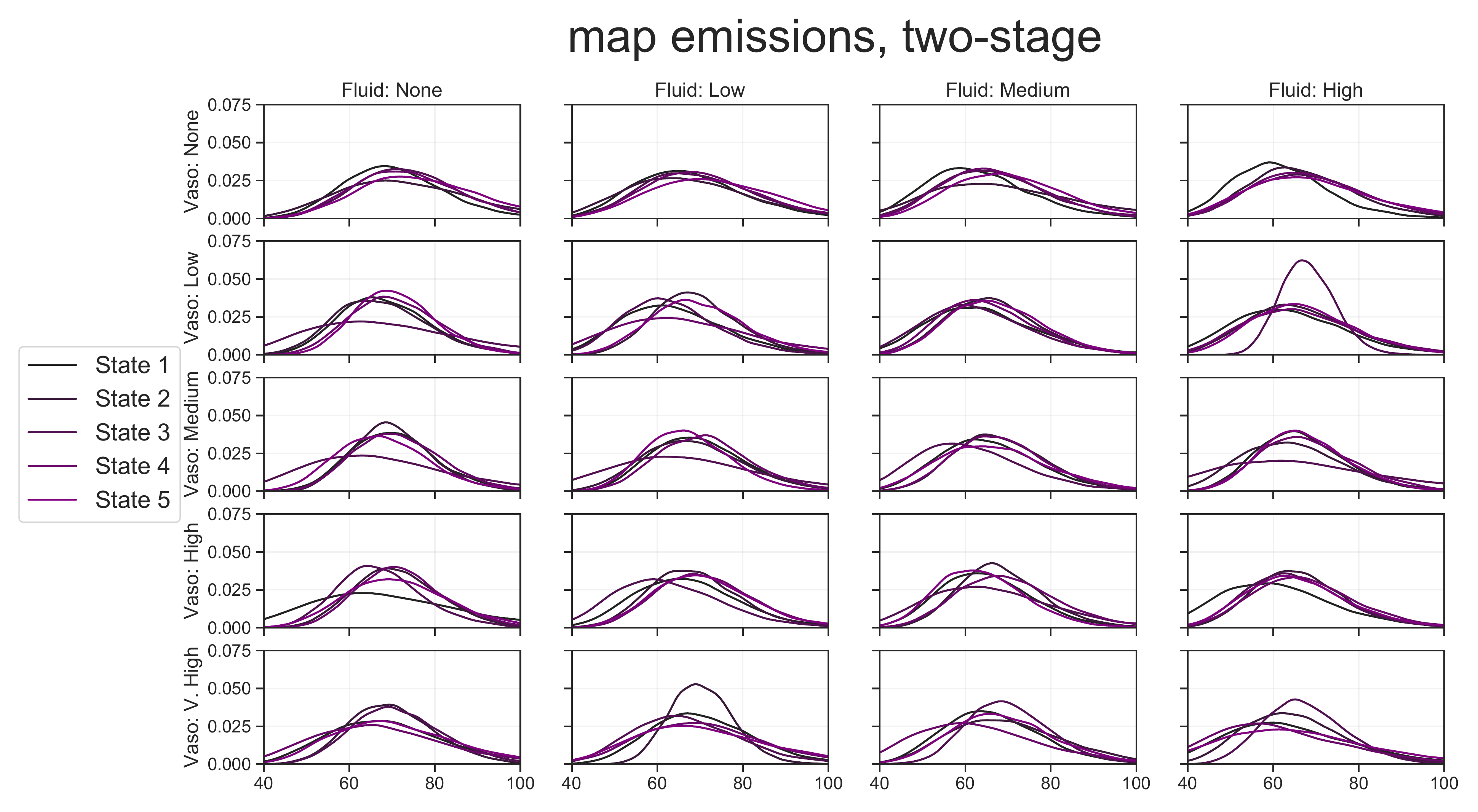}
&
  \includegraphics[height=0.185\textwidth]{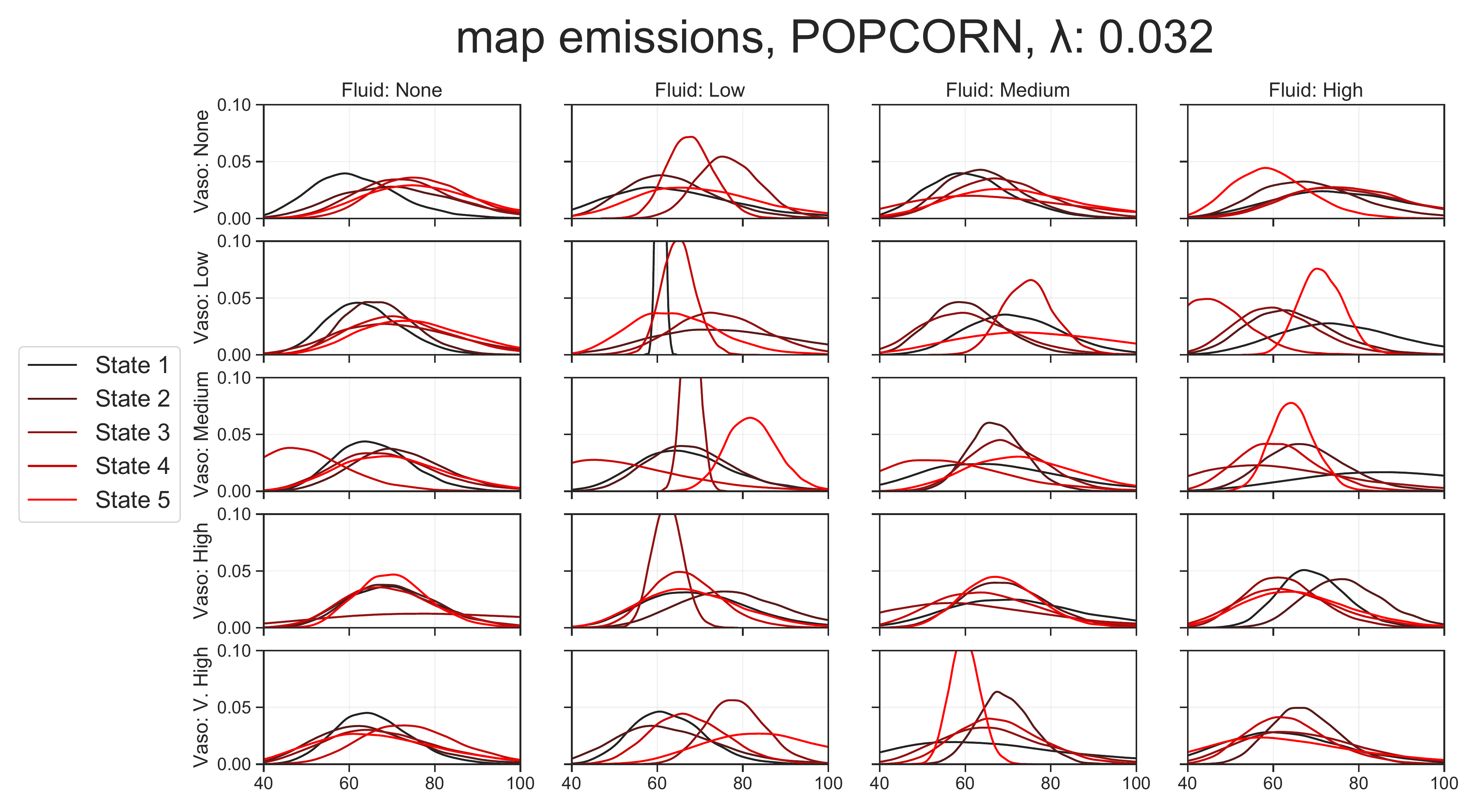}
&
  \includegraphics[height=0.185\textwidth]{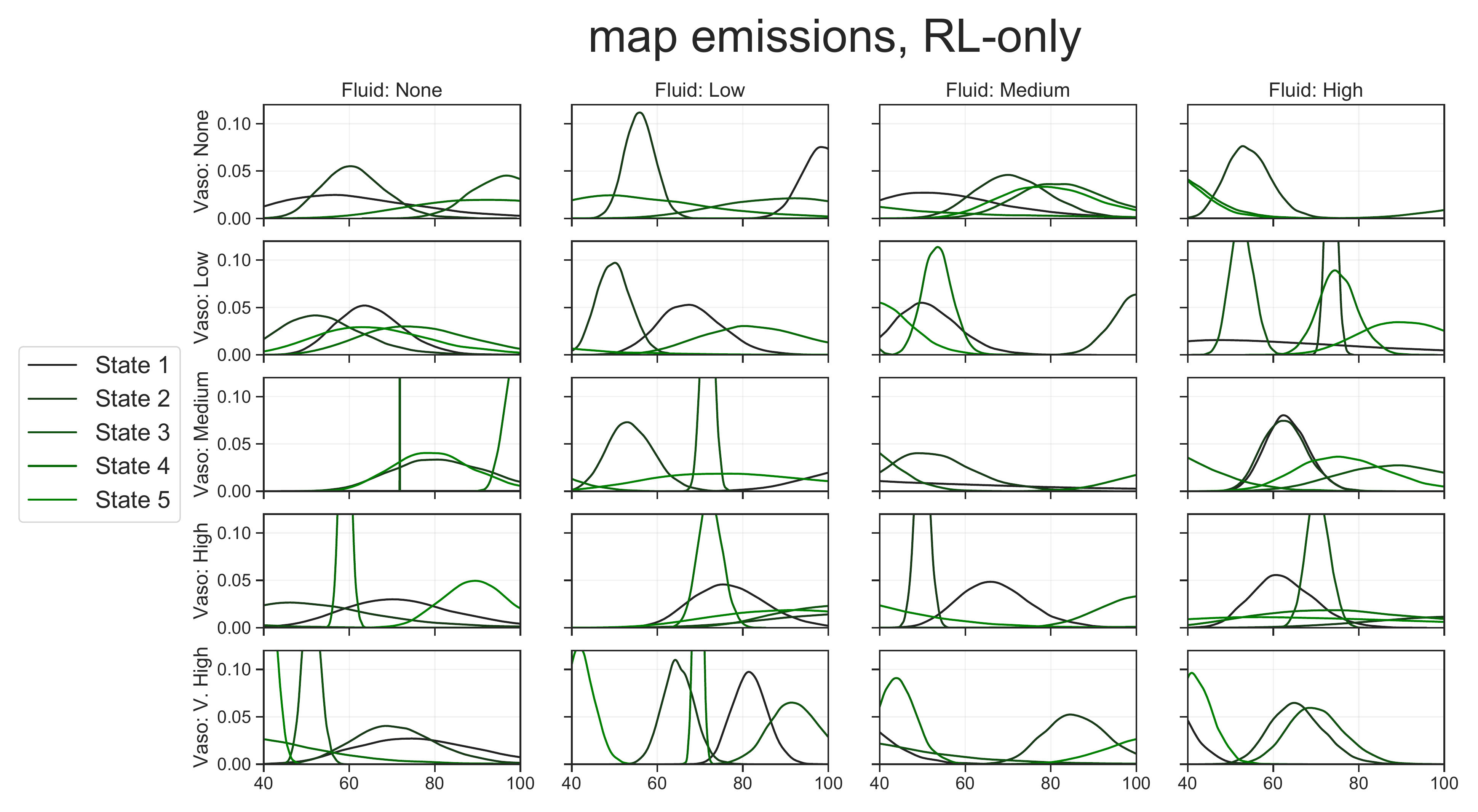}
\end{tabular}
\caption{
Visualization of learned MAP distributions. 
\emph{Left:} 2-stage.
\emph{Middle:} POPCORN, $\lambda=0.032$.
\emph{Right:} Value-only.
Each subplot visualizes all $100$ learned distributions of MAP values for a given method, across $20$ actions and $K=5$ states. Each pane in a subplot corresponds to a different action, and shows distributions across the $5$ states. Vasopressors  vary across rows, and fluids vary across columns. 2-stage learns states that are mostly homogeneous, value-only learns states that are differentiated and often far apart, while POPCORN is somewhere in between.
}
\label{fig:MAP-dists}
\end{figure*}

\begin{figure}[!t]
\centering
\setlength{\tabcolsep}{0.01cm}
\begin{tabular}{c c}
  \includegraphics[height=0.28\textwidth]{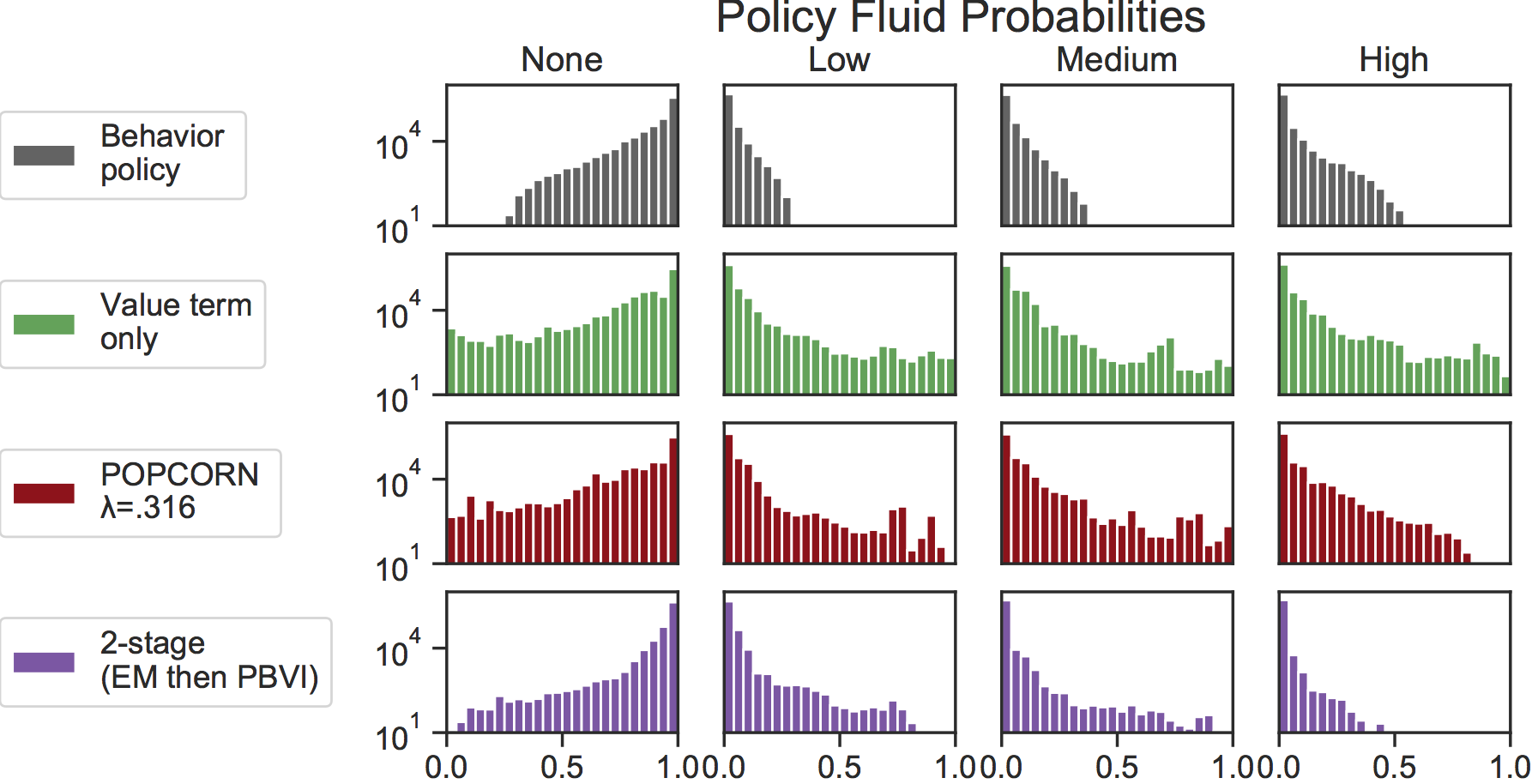}
&
  \includegraphics[height=0.28\textwidth]{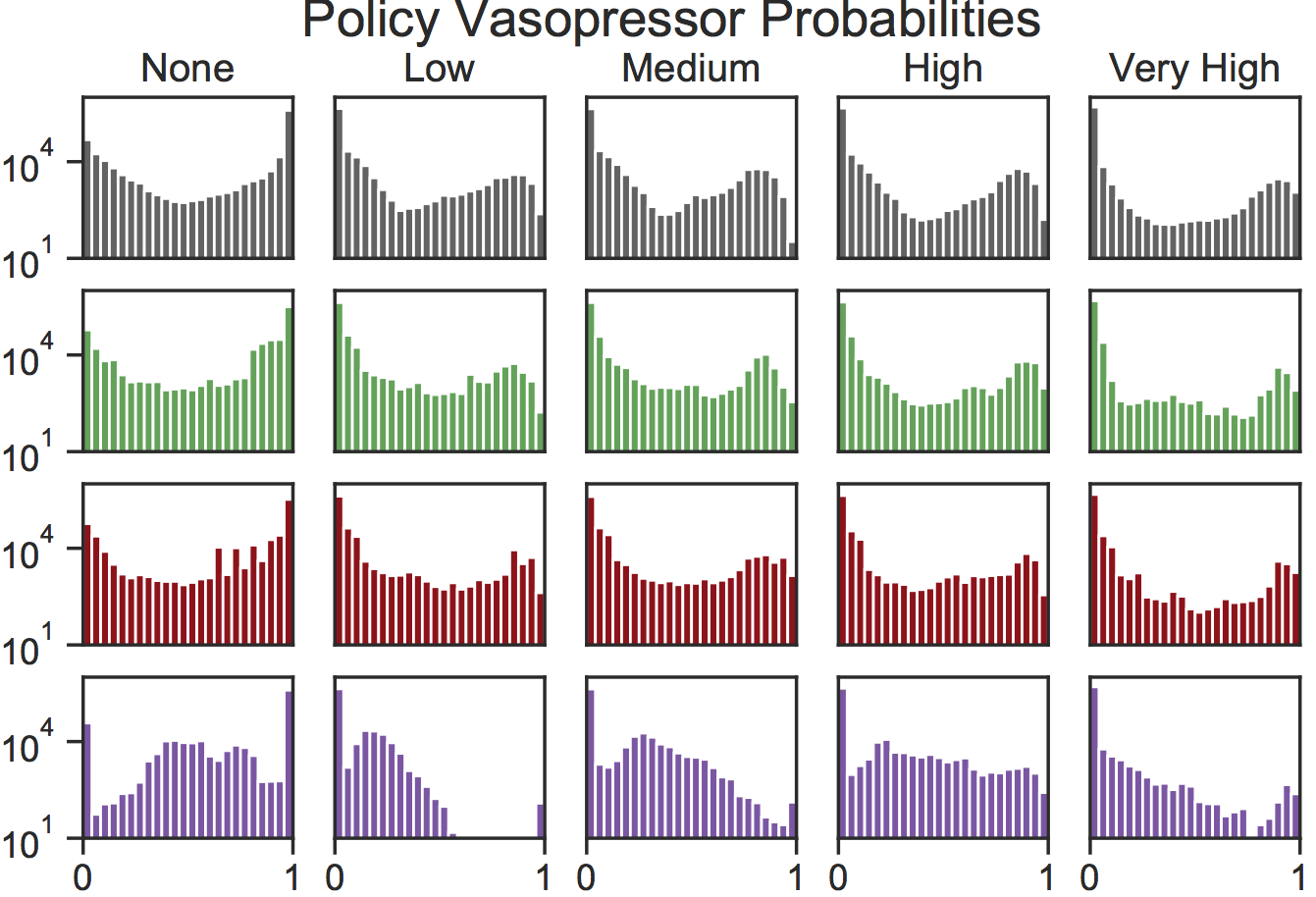}
\end{tabular}
\caption{
Action probabilities for the behavior policy, a value-only policy, a POPCORN policy with $\lambda=0.316$, and a 2-stage policy. Actions are split from the full 20-dimensional space by type. \emph{Left:} Action probabilities for the 4 doses of IV fluids, and \emph{Right:} for the 5 doses of vasopressors.
}
\label{fig:hypo-global-probas}
\end{figure}

\begin{figure}[!t]
\centering
\setlength{\tabcolsep}{0.01cm}
	\includegraphics[width=0.8\textwidth]{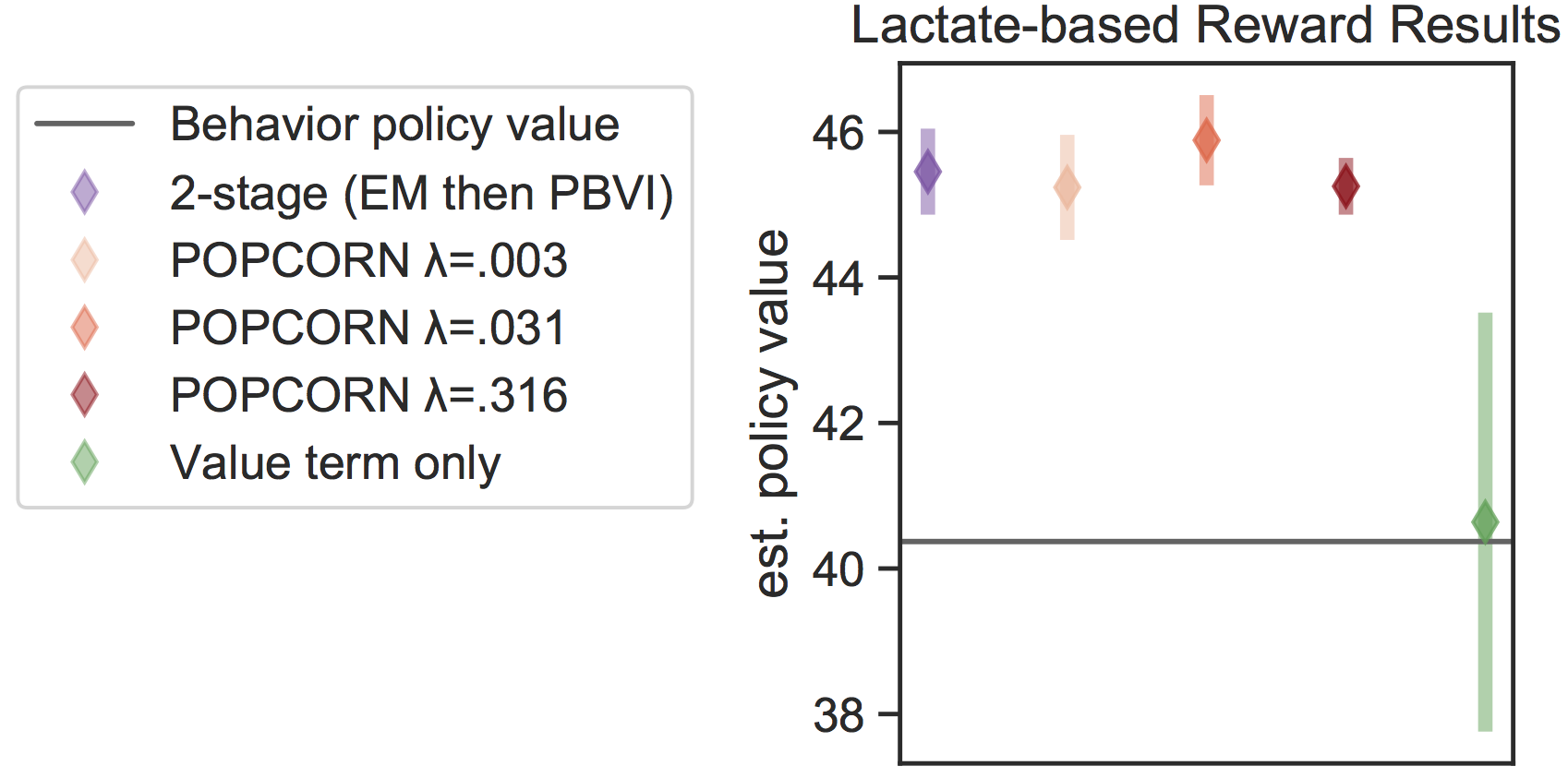}
\caption{
Results from reusing previously learned models from each approach, and solving them to learn new policies given a different reward function based on lactate values. 2-stage and POPCORN learn models that are able to transfer reasonably well to the new task, while value-only does not. 
}
\label{fig:reward-respec}
\end{figure}

\vspace{-0.1in}
\subsection{Conclusions from ICU Application}

\textbf{POPCORN achieves the best balance of high-performing policies and high likelihood models.} 
As in earlier results, Figure \ref{fig:hypo-tradeoff} shows how POPCORN balances generative and decision-making performance well, with darker red indicating higher $\lambda$'s and thus improved policy values. The policy values for the 2-stage baseline and the likelihood scores for the value-only baseline both substantially underperform POPCORN.

\textbf{POPCORN has reasonably accurate forecasts.} To demonstrate the ability of models to predict future observations, Figure~\ref{fig:hypo-forecast} shows results from a forecasting experiment. Each method is given the first 12 hours of a trajectory, and then must predict future observations up to 12 hours in the future. Importantly, only \emph{measured} observations are used to calculate the mean absolute error between model predictions and true values. Unsurprisingly 2-stage generally performs the best, although POPCORN for small values of $\lambda$ often performs similarly. On the other hand, the value-only baseline fares significantly worse. For some observations (MAP and urine output; see left-most column of Figure~\ref{fig:hypo-forecast}), it makes nonsensical predictions far outside the range of observed data, with errors several orders of magnitudes worse than POPCORN and 2-stage.

\textbf{POPCORN enables inspection if learned models are clinically sensible.} We visualize the learned emission distributions for MAP across the $K=5$ states and $20$ actions for each method in Figure~\ref{fig:MAP-dists}. 
Note that densities may appear non-Gaussian, as they are back-transformed to the original scale of the data but were modeled on a log-scale.
POPCORN's distributions are more spread out and better differentiate between states compared to the 2-stage baseline, which learns very similar states with high overlap. 
As a result, the 2-stage policy will end up recommending similar actions for most patients.
Value-only learns states that are even more diverse, allowing it to learn an effective policy but at the expense of not modeling the observed data well.
See Appendix \ref{app:mimic:results} for similar results for lactate, urine output, and heart rate.
Although these results are exploratory, these simple visualizations of what the models have learned are only possible due to the white-box nature of our HMM-based approach, compared with e.g. deep reinforcement learning methods.

Figure~\ref{fig:hypo-global-probas} visualizes the action probabilities for the behavior policy, a value-only policy, a POPCORN policy, and a 2-stage policy.  In general, the POPCORN policy most closely aligns with the behavior, although it is also quite similar to value-only. On the other hand, the 2-stage policy seems in general more conservative and tends to have lower probabilities on more aggressive actions. In future work we plan to work with clinical partners to explore individual patient trajectories and understand how and why these treatment policies differ.

\textbf{POPCORN learns models that transfer to other tasks.} Figure~\ref{fig:reward-respec} shows results testing how well models transfer to solving a new task. We use a new reward function that penalizes high lactate values (see Appendix \ref{app:mimic:rewards} for a plot). For each method, we freeze $\tau, \mu, \sigma$ from the previous optimization, but learn a new $R$. Then we solve these new models to learn new policies, and estimate their values. We find that the POPCORN and two-stage models transfer reasonably well, whereas value-only is substantially worse especially given its high original estimated value in Figure~\ref{fig:hypo-tradeoff}.

\vspace{-0.15in}
\section{Discussion}
\vspace{-0.15in}

We proposed POPCORN, an optimization objective for off-policy batch RL with partial observability. 
POPCORN balances the trade-off between learning a model with high likelihood and a model well-suited for planning, even in batch off-policy settings.
Synthetic experiments demonstrate POPCORN achieves good policies and decent models even in the face of misspecification (in the number of states, the choice of the likelihood, or the availability of data).
Performance on a clinical decision-making task suggests we may be able to learn a policy on par or even slightly better than the observed clinician behavior policy. 
Future directions include scaling to environments with more complex state structures or long-term temporal dependencies, investigating semi-supervised settings where not all sequences have rewards, better leveraging that the behavior policy is not terribly sub-optimal, and learning Pareto-optimal policies that balance multiple competing goals. We hope methods such as ours ultimately become integrated into clinical decision support tools to assist physicians in improving the treatment of critically ill patients.

\subsubsection*{Acknowledgements}
FDV and JF acknowledge support from NSF Project 1750358. JF additionally acknowledges Oracle Labs, a Harvard CRCS fellowship, and a Harvard Embedded EthiCS fellowship. MCH acknowledges support from NSF Project HDR-1934553. The authors also thank David Sontag, Omer Gottesman, Leo Anthony Celi, Ryan Kindle, and the anonymous reviewers for thoughtful and constructive feedback.

\bibliography{references}

\clearpage
\newpage

\appendix

\section{Additional Details on Point-Based Value Iteration}
\label{app:pbvi}

Point based value iteration (PBVI) is an algorithm for efficiently solving POMDPs \citep{Pineau2003}. See \citet{shani2013survey} for a thorough survey of related algorithms and extensions in this area.

\subsection{Background}
\label{app:pbvi:background}

As first observed in \citet{sondik1978pomdp}, the value function for a POMDP can be approximated arbitrarily closely as the upper envelope of a finite set of linear functions of the belief, commonly referred to as $\alpha$-vectors. Representing the value function as a collection of linear functions, we can write the value of an arbitrary belief $b \in \Delta^K$ in the probability simplex as:
\begin{equation}
	V(b) = \max_{\alpha}\; b \cdot \alpha. \label{eq:POMDP-value}
\end{equation}
Each $\alpha$-vector is associated with a corresponding optimal action $a_\alpha$, so the value function can be represented as a set of pairs $\{(\alpha, a_\alpha)\}_{\alpha \in V}$. To act according to this value function representation, the action $a_{\alpha^*}$ associated with $\alpha^*$, the maximizing $\alpha$-vector, is taken given the current belief $b$ at each time point. The task of solving a POMDP is then to compute this set of $\alpha$-vectors. Unfortunately, exactly representing the true value function (via e.g. exact value iteration) requires exponentially many $\alpha$-vectors, and this becomes computationally intractable for even small problems. Early techniques for efficiently solving POMDPs often involved iteratively pruning redundant $\alpha$-vectors at each iteration of the solver, but these approaches also did not scale well. See \citet{shani2013survey} for more details.

\subsection{PBVI Overview}
\label{app:pbvi:overview}

In PBVI, unlike in exact value iteration, we do not perform full Bellman backups over the space of all possible belief points, as this is typically intractable. Instead, we will only perform backups at a fixed set of belief points, which we denote by $\mathcal{B} \triangleq \{b_i\}_{i=1}^B$, with $B = |\mathcal{B}|$ and $b_i \in \Delta^K$. We will return later to how this set is chosen.

We first highlight the computation for the value at a belief $b$ after a Bellman backup over $V$, where we let $r_a$ denote the vector $R(\cdot,a)$:
\begin{equation}
V'(b) = \text{max}_{a \in A} r_a \cdot b + \gamma \sum_o p(o|b,a) V(b^{a,o}),\label{eq:PBVI-value-1}
\end{equation}
where 
\begin{equation}
	\alpha^{a,o}(s) = \sum_{s'} \alpha(s') p(o|s',a) p(s' | s,a), \label{eq:alpha-ao}
\end{equation}
and $b^{a,o}(s') = p(s' | b,a,o)$ denotes our new belief that we are in state $s'$ having started from belief vector $b$, taken $a$, and seen $o$. 

See \citet{shani2013survey} for the full derivation, but some algebra eventually reduces this expression to:
\begin{equation}
V'(b) = \text{max}_{a \in A} r_a \cdot b + \gamma \sum_o \text{max}_{\alpha \in V} b \cdot \alpha^{a,o}. \label{eq:PBVI-value-2}
\end{equation}
The two maxes in this equation implicitly prunes dominated $\alpha$-vectors twice, which is more efficient than previous approaches that would first enumerate the (massive) space of all $\alpha$-vectors and then prune afterwards.

We can use the value function computation in Eq.\ref{eq:PBVI-value-2} to efficiently compute the new $\alpha$-vector that would have been optimal for $b$, had we ran the complete Bellman backup:
\begin{align}
backup(V,b) &= \text{argmax}_{\alpha_a^b: a \in A, \alpha \in V} b \cdot \alpha_a^b \label{eq:PBVI-pt-backup-1} \\
\alpha_a^b &= r_a + \gamma \sum_o \text{argmax}_{\alpha^{a,o}: \alpha \in V} b \cdot \alpha^{a,o} \label{eq:PBVI-pt-backup-2}
\end{align}
During the backup, the action associated with the new $\alpha$-vector is also cached. Importantly, these point-based updates are substantially more efficient than an exact update, as they are quadratic rather than exponential. In addition, for problems with finite horizons, the error between the PBVI approximate value function and the true value function decreases to 0 as we more densely sample the belief simplex and we take $B \to \infty$. 

\paragraph{Choose the Belief Points.} We now briefly discuss how the set $\mathcal{B}$ is chosen. There are many different implementation choices that can be made; see \citet{shani2013survey} for a comprehensive list of previous works of approaches made in different algorithms. A naive approach is to randomly the simplex or choose beliefs evenly spaced on a grid, but both are usually inefficient and may include many beliefs that, in practice, would rarely be reached by actual trajectories. 

We use the strategy used in the original PBVI paper \citep{Pineau2003}. Start with an initial set of beliefs $\mathcal{B}_0$. In our work, we initialize this to be the uniform belief vector $\frac{1}{K}\vec{1}$ along with beliefs that place a large amount of mass (e.g. 99\%) on a single state. In the end, our initial set $\mathcal{B}_0$ contains $K+1$ belief points. To add a new belief point $b$ to an existing set $\mathcal{B}$, we find a successor belief $b'$ that is most distant from our current set. We do this by using a distance metric (in practice, we use standard $L_2$ Euclidean distance), and let
\begin{equation}
	|b' - \mathcal{B}| = \text{min}_{b \in \mathcal{B}} |b - b'|
\end{equation}
be the distance from a new belief $b'$ to the set $\mathcal{B}$. We focus on new candidate beliefs that can be reached from the current set. For discrete observations, we can enumerate all possible $b^{a,o}$ that are reachable given a starting belief $b$, if we take each possible action $a$ and then see observation $o$. For continuous observations, we of course cannot enumerate all possible $b^{a,o}$ but can instead just draw samples from our observation model. We can then add to $\mathcal{B}$ an additional new belief $b'$ that is farthest from $\mathcal{B}$. Or, we can add a set of new beliefs that are all ``far'' from the current set $\mathcal{B}$, e.g. greater than some pre-specified distance $\epsilon$.  The  high-level idea for this belief set expansion is for the set $\mathcal{B}$ to be spread out relatively evenly across the \emph{reachable} parts of belief space. 

\paragraph{Implementation Details.} In practice, we may interleave belief expansion steps where we increase the size of $\mathcal{B}$ with a large number of backups, repeatedly running Eq~\ref{eq:PBVI-pt-backup-1} for all current beliefs in $\mathcal{B}$. 

Having the entire PBVI algorithm as a subroutine in the larger optimization pipeline for POPCORN is a challenging task. One rather expensive and inefficient design choice would be to entirely rerun PBVI, \emph{from scratch}, at each iteration of gradient descent. Instead, we cache the intermediate value function and the current set of belief points during optimization. During one gradient update, we then choose to only run a small number of PBVI backups (in practice, between 1-5), where we run backups of our current beliefs given the new model parameters $\theta$ at this iteration of training.

As we are learning both the policy and the model online during training for POPCORN, we empirically found that it is helpful to occasionally do a hard reset of both $\mathcal{B}$ and our value function. In the planning community (e.g. the original PBVI paper), it is typically assumed that the ground truth model $\theta$ is known, whereas in real-world settings, the model must be learned from data. This means that during training, our estimate of $\theta$ constantly changes, and over time our value function and belief set may have been largely determined from very stale previous values of $\theta$. In practice, we do these hard resets very infrequently, e.g. only once every 250 or 500 gradient updates.

\subsection{PBVI: Sampling Approximation to Deal with Complex Observation Models}
\label{app:pbvi:cts-obs}

In normal PBVI, we are limited by how complex our observation space is. The PBVI backup crucially depends on a summation over observation space (or integration, for continuous observations).  Dealing with multi-dimensional, non-discrete observations is generally intractable to compute exactly. 

Instead, we will utilize ideas from \citet{Hoey2005} to circumvent this issue. The main idea is to learn a partition of observation space, where we group together various observations that, conditional on a given belief $b$ and taking an action $a$, would have the same maximizing $\alpha$-vector.  That is, we want to learn $\mathcal{O}_\alpha = \{o | v = \text{argmax}_{\alpha \in V} \alpha \cdot b^{a,o} \}$. We can then treat this collection of $\mathcal{O}_\alpha$ as a set of ``meta-observations'', which will allow us to replace the intractable sum/integral over observation space into a sum over the number of $\alpha$-vectors, by swapping out the $p(o | b,a)$ term in Equation \ref{eq:PBVI-value-1} with $p(\mathcal{O}_\alpha | b,a)$, the (approximate) aggregate probability mass over all observations in the ``meta-observation''. In particular, we can express the value of a belief by:
\begin{align}
V(b) &= \text{max}_a r_a \cdot b + \gamma \sum_{\alpha} p(\mathcal{O}_{\alpha} | b,a) V(b^{a,\mathcal{O}_\alpha}) \\
p(\mathcal{O}_{\alpha} | b,a) &= \sum_s b(s) \sum_{s'} p(s'|a,s) p(\mathcal{O}_{\alpha} | s',a) \\
b^{a,\mathcal{O}_\alpha} &\propto p(\mathcal{O}_{\alpha} | a,s') \sum_s b(s) p(s'|s,a) \\
p(\mathcal{O}_{\alpha} | a,s') &= \sum_{o \in \mathcal{O}_{\alpha}} p(o|a,s').
\end{align}
We will make use of a sampling approximation that admits arbitrary observation functions in order to approximate the $\mathcal{O}_\alpha$ and $p(\mathcal{O}_{\alpha} | s',a)$, the aggregate probability of each ``meta-observation''.

To do this, first we sample $k$ observations $o_k \sim p(o|s',a)$, for each pair of states and actions. Then, we can approximate $p(\mathcal{O}_{\alpha} | a,s')$ by the fraction of sampled $o_k$ where $\alpha$ was the optimal $\alpha$-vector, ie 
\begin{equation}
p(\mathcal{O}_{\alpha} | a,s') \approx \frac{| \{o_k : \alpha = \text{argmax}_{\alpha \in V} \alpha \cdot b^{a,o_k} \} |}{k}, \label{eq:prob-meta-obs}
\end{equation}
  where ties are broken by e.g. favoring the $\alpha$-vector with lowest index.  Using this approximate discrete observation function, we can perform point-based backups for $V$ at a set of beliefs $\mathcal{B}$ as before. Our backup operation is now:
\begin{align}
backup(V,b) &= \text{argmax}_{\alpha_a^b: a \in A, \alpha \in V} b \cdot \alpha_a^b \\
\alpha_a^b &= r_a + \gamma \sum_{\alpha'} \underset{\alpha^{a,\mathcal{O}_{\alpha'}}}{\text{argmax }} b \cdot \alpha^{a,\mathcal{O}_{\alpha'}} \\
\alpha^{a,\mathcal{O}_{\alpha'}}(s) &= \sum_{s'} \alpha(s') p(s'|s,a) p(\mathcal{O}_{\alpha'}|a,s').
\end{align}
The previous sum/integral over observations has now been replaced by a sum over $\alpha$-vectors, which is generally more tractable.

\subsection{Softmax Relaxation to Make PBVI Differentiable}
\label{app:pbvi:softmax-relax}

In order to be able to differentiate through the entire PBVI backups and allow gradient-based optimization for POPCORN, we relax the original argmax operations involved in PBVI backups and running a PBVI policy to softmaxes. There are 2 argmax operations in the original PBVI backups, in Eqs.~\ref{eq:PBVI-pt-backup-1} and \ref{eq:PBVI-pt-backup-2}. For PBVI with continuous observations, there is an additional argmax associated with the probability of ``meta-observations'' in Eq.~\ref{eq:prob-meta-obs}. Lastly, there is a fourth argmax associated with actually running a policy, as we need to determine which $\alpha$-vector is the maximizing one, and we take its corresponding action.

We relax all 4 of these argmaxes to softmaxes. Previously, there were operations such as $\text{argmax}_i \; x_i$ to select a maximal index; we can view these as returning a delta function at the maximizing index, or a probability mass function with all mass on one element. We instead relax this to a softmax, now returning $p \triangleq \frac{e^{x_i}}{\sum_i e^{x_i}}$, a distribution over all elements. Where before we might have taken, e.g., $\alpha_j$ if $j$ was the maximizing index of $x$, now we instead take a \emph{soft} mean using the softmax probability distribution $p$, i.e. we instead would take $A \cdot p$ where $A \in \mathbb{R}^{K \times N}$ is a matrix with all $N$ vectors $\alpha \in \mathbb{R}^K$ stacked up, and $p$ is a probability vector over the $N$ choices. 

Last, we further modified these softmaxes by using an additional temperature parameter $\tau$, which lets us control how close to deterministic the softmax is. That is, we redefine the softmax as $p \triangleq \frac{e^{x_i/\tau}}{\sum_i e^{x_i/\tau}}$. As $\tau \to 0$, the softmax $p$ approaches the deterministic argmax, while $\tau \to \infty$ approaches a uniform distribution. In experiments, throughout we used a fixed $\tau=0.01$ for all environments. In initial tests on the tiger environment, we tried starting with larger temperatures and slowly annealing them to smaller values, but found this only added noise and slowed overall convergence. For relatively small temperatures, we confirmed that the softmax-relaxed PBVI solutions were comparable to the original deterministic ones. 

Note that in this relaxation, each $\alpha$-vector is now associated with a \emph{distribution} over actions, rather than a single action as before. Likewise, as we now learn stochastic policies, to run a soft-PBVI policy, we take a soft mean of the softmax distribution over actions associated with each $\alpha$-vector; contrast this with the deterministic solution where we'd simply choose the action associated with the maximizing $\alpha$-vector. 

It is also worth noting that in simulated environments where we can actually run a policy, we can always run a deterministic version of a softmax policy by simply selecting the most likely action, rather than probabilistically choosing an action from a policy's distribution over actions. Our main motivation for using softmax policies is that it makes OPE easier, as otherwise for deterministic policies we may run into severe problems if the support of our deterministic policy has little in common with the behavior policy.

\section{Additional Details on Learning Rewards}
\label{app:learning-rewards}

We noted in the main text that learning rewards is explicitly not part of the main optimization procedure in POPCORN. This is because we expressly do \emph{not} want to compute gradients for the estimated policy value term with respect to the reward function parameters. If we did so, there would be nothing stopping the optimization procedure to ``hallucinate'' that the best way to learn a high-value policy is to simply increase all values of the reward function to be large. The policy induced by such a model with incorrectly high rewards would then appear to be very good, with respect to the model, but when run in the real world or real environment, it would perform terribly.

Instead, we simply learn the rewards a separate EM step, that may be performed alongside each gradient update to $\tau, \mu, \sigma$ (in practice this is what we do), or may be done only periodically. In the E-step, we compute the relevant summary statistics from the forward and backward pass through the IO-HMM. Importantly, the E-step does \emph{not} depend on the reward function $R$ at all; the forward and backward pass only use the transition and emission distributions to update relevant state probabilities. Then, in the M-step, we update \emph{only} the reward function, using the observed reward values from the trajectories in our dataset. This is equivalent to minimizing the sum of squared errors between our reward function and the observed reward values. It is also nearly identical to what the M-step looks like for $\mu$, but the other observations are assumed to be Gaussian and hence also have the variance $\sigma$ parameters. 

\section{Additional Setup Details and Results for Tiger Domains}
\label{app:tiger}

We show a few extra results from the synthetic tiger domains, and provide a bit more detail for the setup for the third environment involving misspecification in the emission distribution itself.

\begin{figure}[h!]
\centering{
\includegraphics[width=\textwidth]{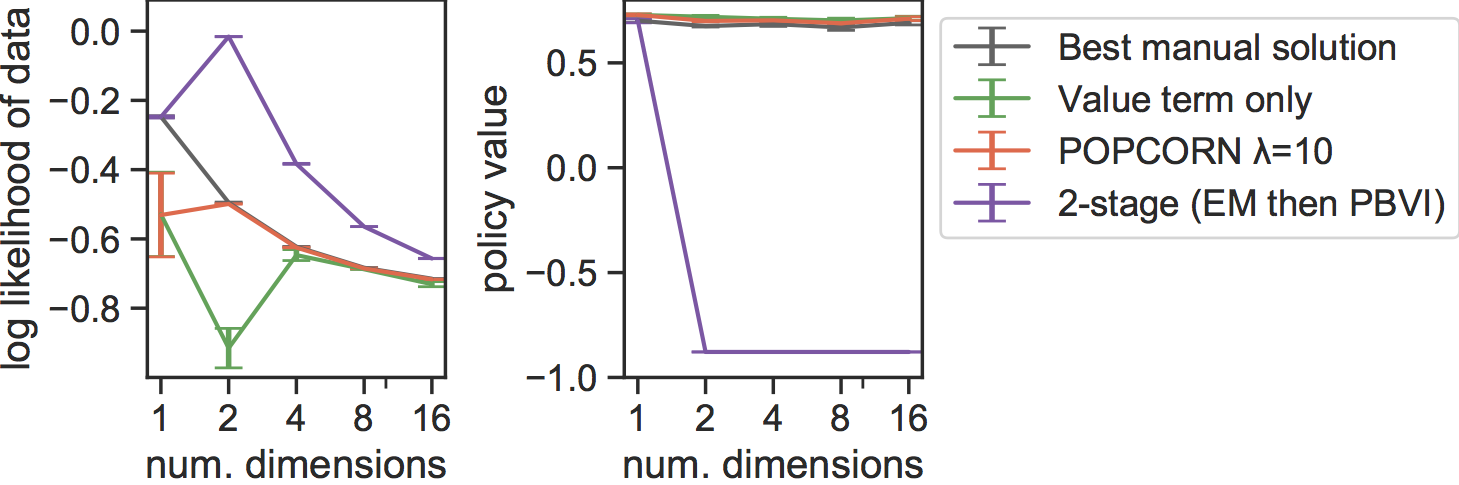}
}
\caption{Results from the Tiger with Irrelevant Noise environment, where we now vary the overall observation dimension size. Throughout, the first dimension is the only relevant dimension with any information about the decision-making task ($\sigma=0.3$ for this dimension) while the rest all contain irrelevant observations with lower $\sigma=0.1$.}
\label{fig:tiger-vary-num-dims}
\end{figure}

First, in Figure \ref{fig:tiger-vary-num-dims} we show results from the first Tiger with Irrelevant Noise environment, where we vary the total dimensionality of observation space from 1 (model is properly specified) to 16, with the results in the main paper only showing 2. Throughout, 2-stage always learns models with the highest likelihood but fails completely at the downstream decision-making task. The difference in likelihood between POPCORN and value-only becomes more muted for larger numbers of dimensions, as the models are more and more misspecified, and the likelihood metric collapses over all dimensions. The differences would remain more apparent if we instead showed the associated likelihood metric for each individual dimension of observation space.

Next, in Figure \ref{fig:tiger-vary-missingness} we show results for the Tiger with Missing Data environment as a function of the fraction of missingness in the relevant dimension needed to make decisions. We vary this amount in the following range: $\{10\%, 30\%, 50\%, 70\%, 80\%, 90\%, 95\%\}$. As the overall amount of missingness increases, likelihood values and policy values generally degrade, which is to be expected as less and less total information is contained in a single dataset. Notably, 2-stage always learns much better models but performs terribly in its policy, while the converse is true for value-only.

\begin{figure}[h!]
\centering{
\includegraphics[width=\textwidth]{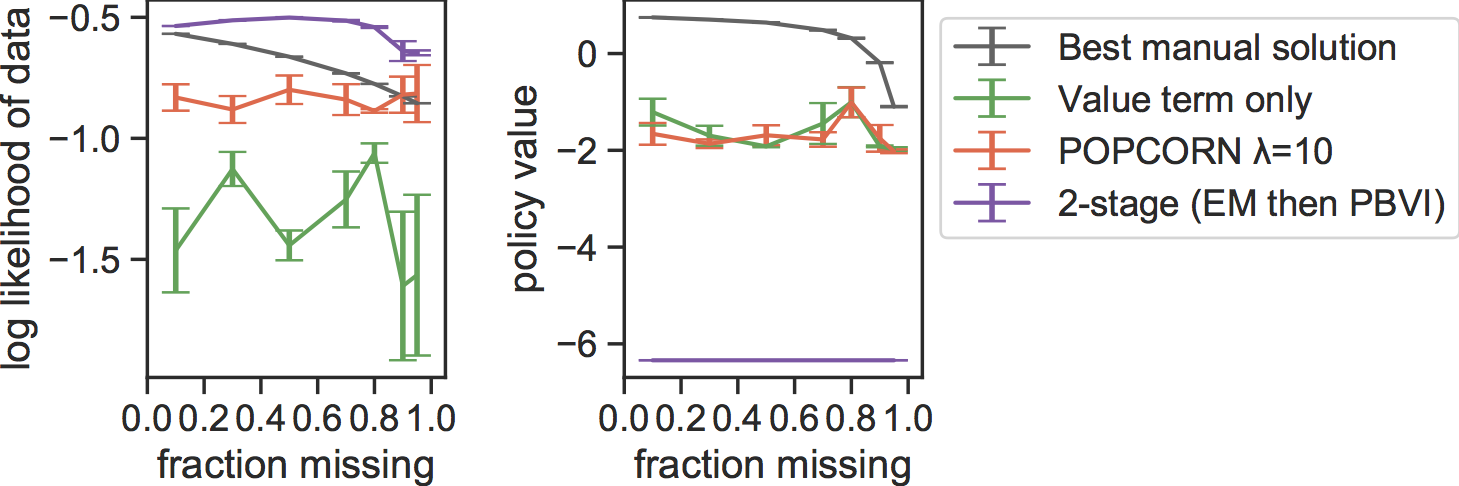}
}
\caption{Additional results from the Tiger with Missing Data. These results show how performance varies as a function of the amount of missing data. In the main text, we only show results from the case where 80\% of the relevant observation variable is missing.}
\label{fig:tiger-vary-missingness}
\end{figure}

Last, we provide qualitative probing of the models learned in the last tiger environment, Tiger with Wrong Likelihood. The true emission distribution is given by a truncated Gaussian Mixture Model (GMM) with equal weights, and the truncation depends on the true unknown latent state value. However, we have setup the emissions by choosing an appropriate prior distribution over the latent states so that marginally, the observations look like they come from a normal GMM and so a pure likelihood-based approach would try to fit a GMM rather than the true truncated GMM. See Figure \ref{fig:tiger-vary-missingness} for the models from the manual solution, 2-stage, and POPCORN. The manual solution used oracle knowledge of the true underlying states which other methods did not have access to, and simply moment matched by taking empirical means and variances. Since the results of value-only and POPCORN were near identical for this environment, we do not show it. For this environment, POPCORN does even better than the manual solution. While 2-stage learns a slightly high likelihood model, its policy is substantially worse. 

In the figure, the histograms show observed observations in green and blue, and model densities in purple and red. The green and blue histograms show observation values colored by their true state. Green bars correspond to state ``1'', so that observations are drawn from the GMM but truncated to be positive. Likewise, blue bars correspond to state ``0'', and observations from this state are drawn from the GMM and then truncated to be negative. The numbers in each subplot denote the learned mean and variance parameter for the 2 states for the emission model for each method (conditioned on the last action being listen). Note that the ground truth means of the truncated GMM were 0 and 1, and the ground truth standard deviations were 0.1 and 1. 2-stage correctly recovers these, but it is tricked into learning a GMM, rather than the true underlying \emph{truncated} GMM. Histograms of the 2 emission distributions learned by each approach are shown in red and purple. 2-stage learns the true parameters of the GMM, whereas POPCORN learns state emissions that are more spread out so that it can correctly differentiate between the two true underlying states. The true underlying states can be perfectly identified by whether or not they are positive or negative; which of the two GMM mixture components they came from does not always identify them correctly.

\begin{figure}[h!]
\centering{
\includegraphics[width=\textwidth]{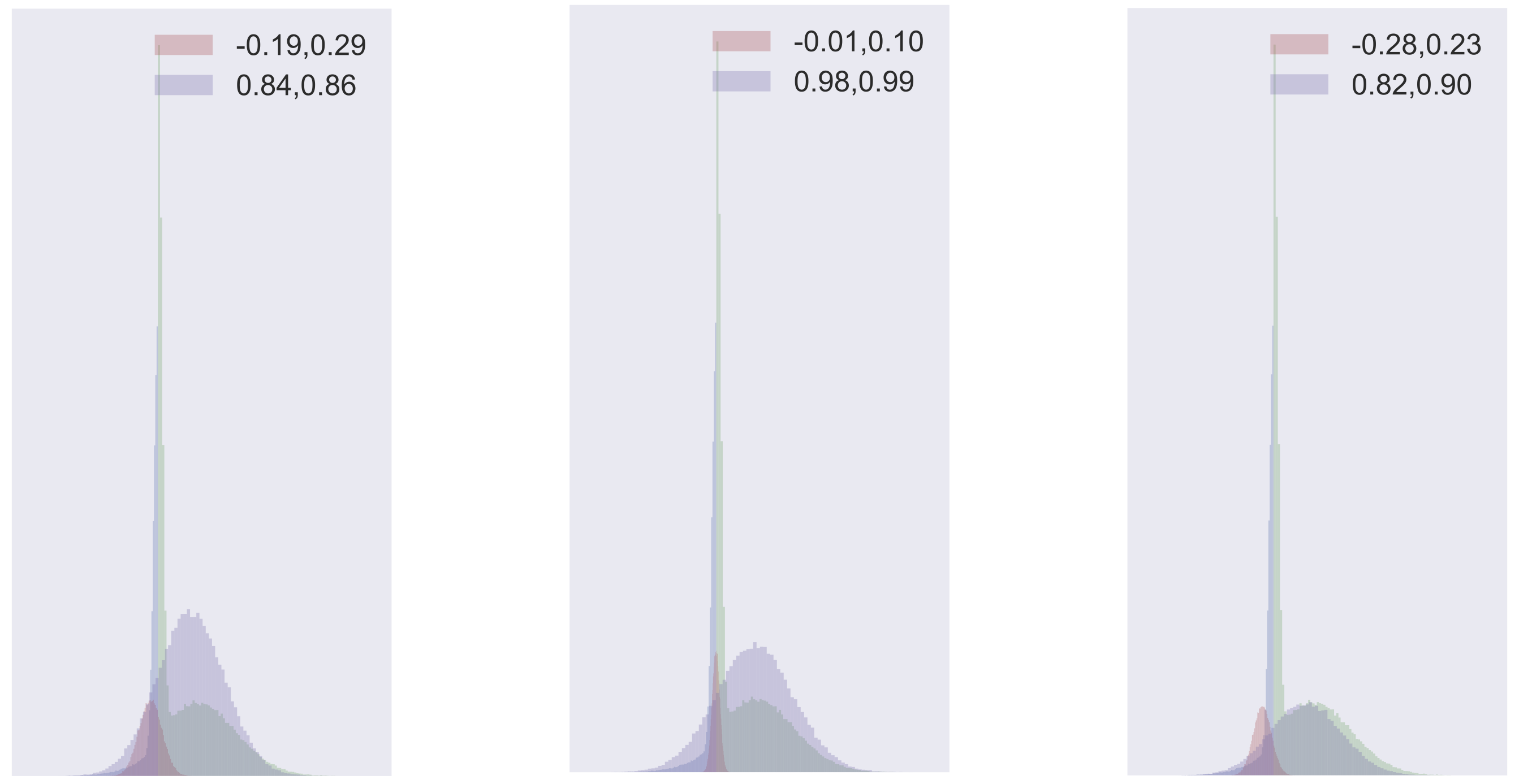}
}
\caption{Qualitative results from the Tiger with Wrong Likelihood environment. \emph{Left}: Manual solution, log marginal likelihood $-0.95$, policy value $0.20$. \emph{Middle}: 2-stage EM solution, log marginal likelihood $-0.76$, policy value $-0.57$. \emph{Right}: POPCORN with $\lambda=1$, log marginal likelihood $-0.92$, policy value $0.50$. See text for details.}
\label{fig:tiger-vary-missingness}
\end{figure}

\section{Additional Setup Details for Sepsis Domain}
\label{app:sepsis}

The original sepsis environment in \cite{oberstCounterfactualOffPolicyEvaluation2019} consists of $D=5$ discrete observation dimensions. Four are vitals and naturally ordinal (e.g. ``low'', ``high''), while the last is binary.

We encode each ordinal discrete observation with $C$ categories as an integer in $\{0,\dots,C-1\}$, then add independent Gaussian noise to this integer, with $\sigma=0.3$. Adding noise to the 4 ordinal-valued vitals is reasonable and can be viewed as approximating measurement error, in some sense, if we pretend that the original discrete variables were obtained by dichotomizing some ``true'' unknown continuous value. This is not strictly true in practice, as the environment is hard-coded and not actually based on some sort of underlying dynamical system. However, adding noise to the diabetes indicator is just a convenient way to make it continuous-valued.

As noted in the main text, we use this environment simply as a slightly more challenging medically-inspired simulator. This differs substantially from its original use in \cite{oberstCounterfactualOffPolicyEvaluation2019} where they used it to introduce strong hidden confounding with known structure by masking 2 state variables from their methods.

In our work we used $K=5$ somewhat arbitrarily. The main purpose of this environment was to illustrate the tradeoff POPCORN makes between likelihood and policy value, and not to try to actually solve the environment or learn a policy that is near-optimal. Given the partially observed nature of our alteration to the environment, and the high noise level with our $\epsilon$-greedy behavior environment, it's not immediately obvious what the best achievable policy that can be learned in the batch setting is. This will be less than the value of the true optimal policy for the original MDP, which is what we showed in the results figure in the main text.

\section{Additional Setup Details and Results from MIMIC ICU Hypotension Domain}
\label{app:mimic}

\subsection{Data Preprocessing}
\label{app:mimic:data}
We did very little filtering to the initial raw dataset. We started with only patients who had data from the MIMIC-III MetaVision database, as this more recent data has better metadata around treatment timing. We only filtered by the first ICU stay for hospital stays that had multiple ICU admits, and then filtered to ICU stays with 3 or more MAP measurements less than 65mmHg. To discretize time, we started 1 hour into ICU admit and used time points at hour 2, etc. up until hour 72, so that at most our trajectories contain 71 actions. We leave to future work to come up with improved, potentially data-driven methods for more realistic time discretization. 

Since an IO-HMM generative model can easily handle missing data, we do not impute missing values for time points when a measurement is missing. In the event that multiple measurements were taken in the span of a single hour, we take the most recent value. This is extremely uncommon for lab values, and only really applied to vitals such as MAP or heart rate. Even then, in MIMIC-III most of the time vitals are logged only once an hour.

Before modeling the physiological values, we log transformed them (after adding 1 to avoid numeric issues). After log transforms, the population distributions for each variable looked reasonably close to normally distributed. This step was necessary as many clinical values have a heavy right tail, and would be inappropriately to model with Gaussians. After the log-transform, we further standardized each variable to have zero mean and unit variance.

\subsection{Action Space Construction}
\label{app:mimic:actions}

\begin{figure}[h!]
\centering{
\includegraphics[width=\textwidth]{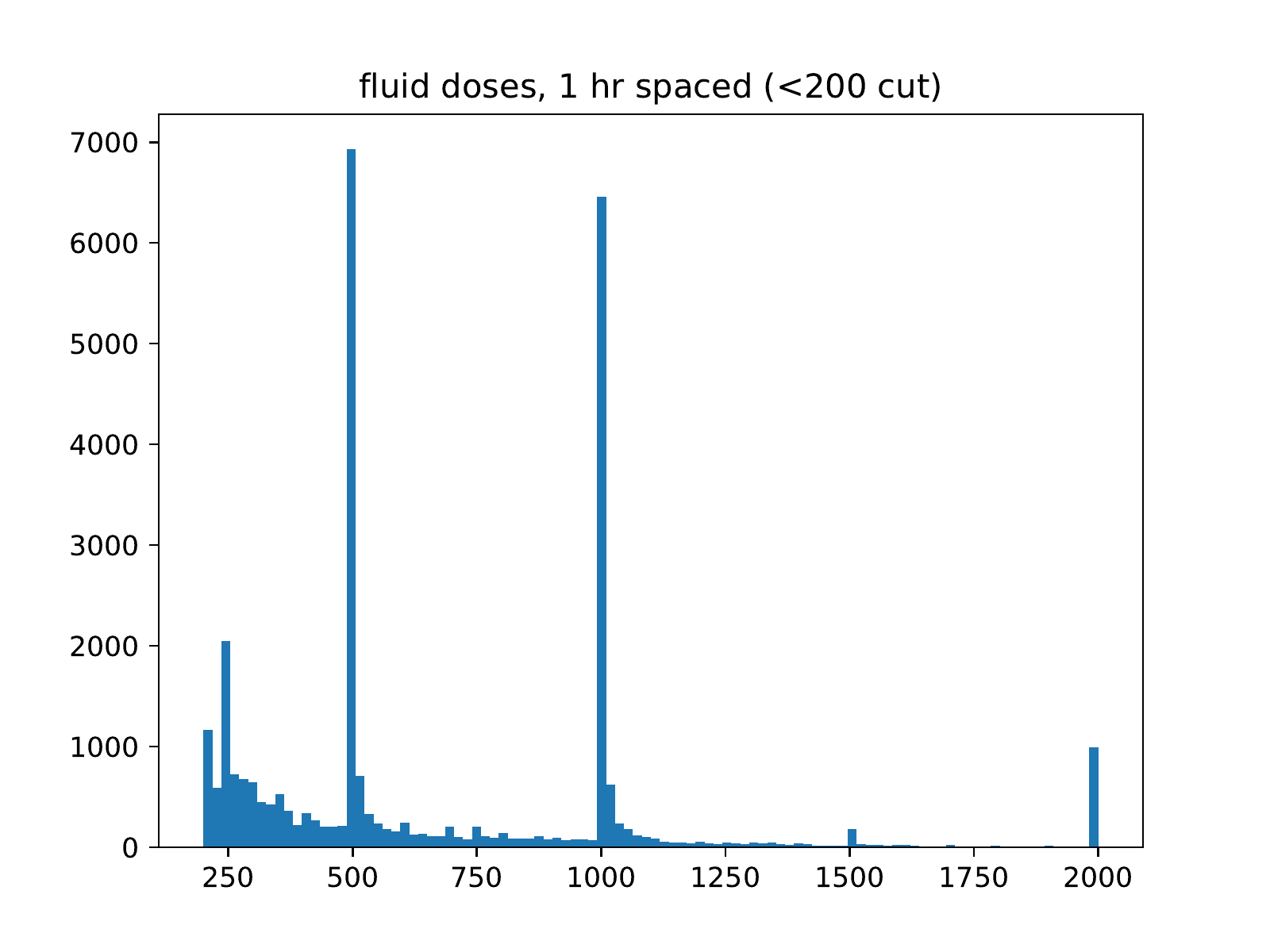}
}
\caption{Distribution of raw IV Fluid doses in the original dataset prior to discretization, in mL.}
\label{fig:data-IV}
\end{figure}

IV fluids somewhat naturally cluster into discrete bins, so this action variable was easier to discretize. We used 4 bins by amount: $\{0, [200,500), [500,1000), [1000,2000]\}$. Figure~\ref{fig:data-IV} shows the distribution of raw fluid amounts in the data.

\begin{figure}[h!]
\centering{
\includegraphics[width=\textwidth]{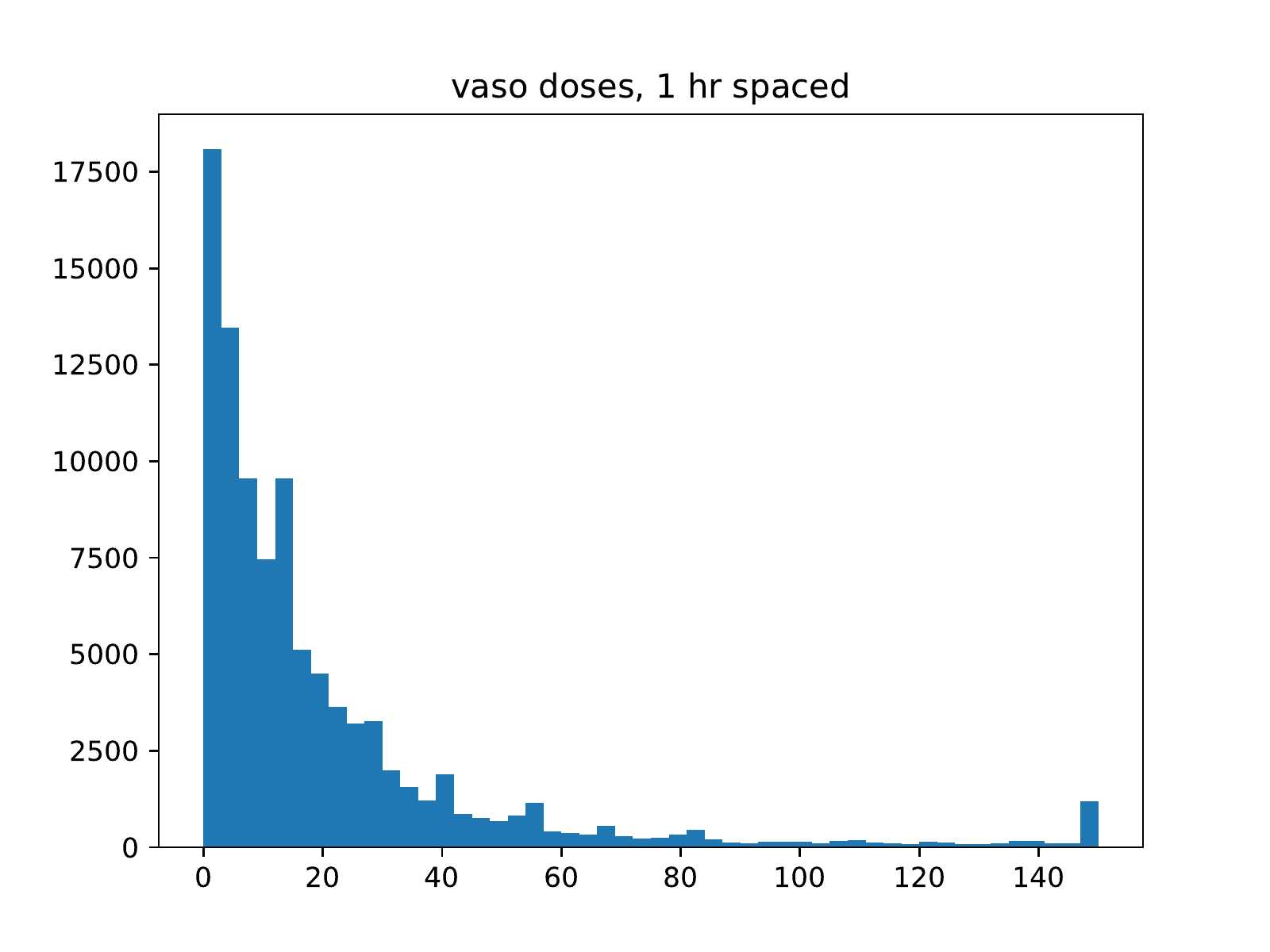}
}
\caption{Distribution of raw vasopressor amounts administered per hour in the original dataset prior to discretization, in (normalized) mcg/kg/hr.}
\label{fig:data-vaso}
\end{figure}

We normalized across vasopressor drug types following the logic in \citet{komorowski2018artificial} in order to arrive at equivalent rates across drugs. Then we took the total amount of vasopressor administered within each hour-long decision window given our time discretization. We eventually then grouped into 5 bins by total drug amount given each hour: $\{0, (0,5), [5,15), [15,40), [40,150]\}$ with units of mcg/kg/hr. Figure~\ref{fig:data-vaso} shows the distribution of raw fluid amounts in the data.

\begin{figure}[h!]
\centering{
\includegraphics[width=\textwidth]{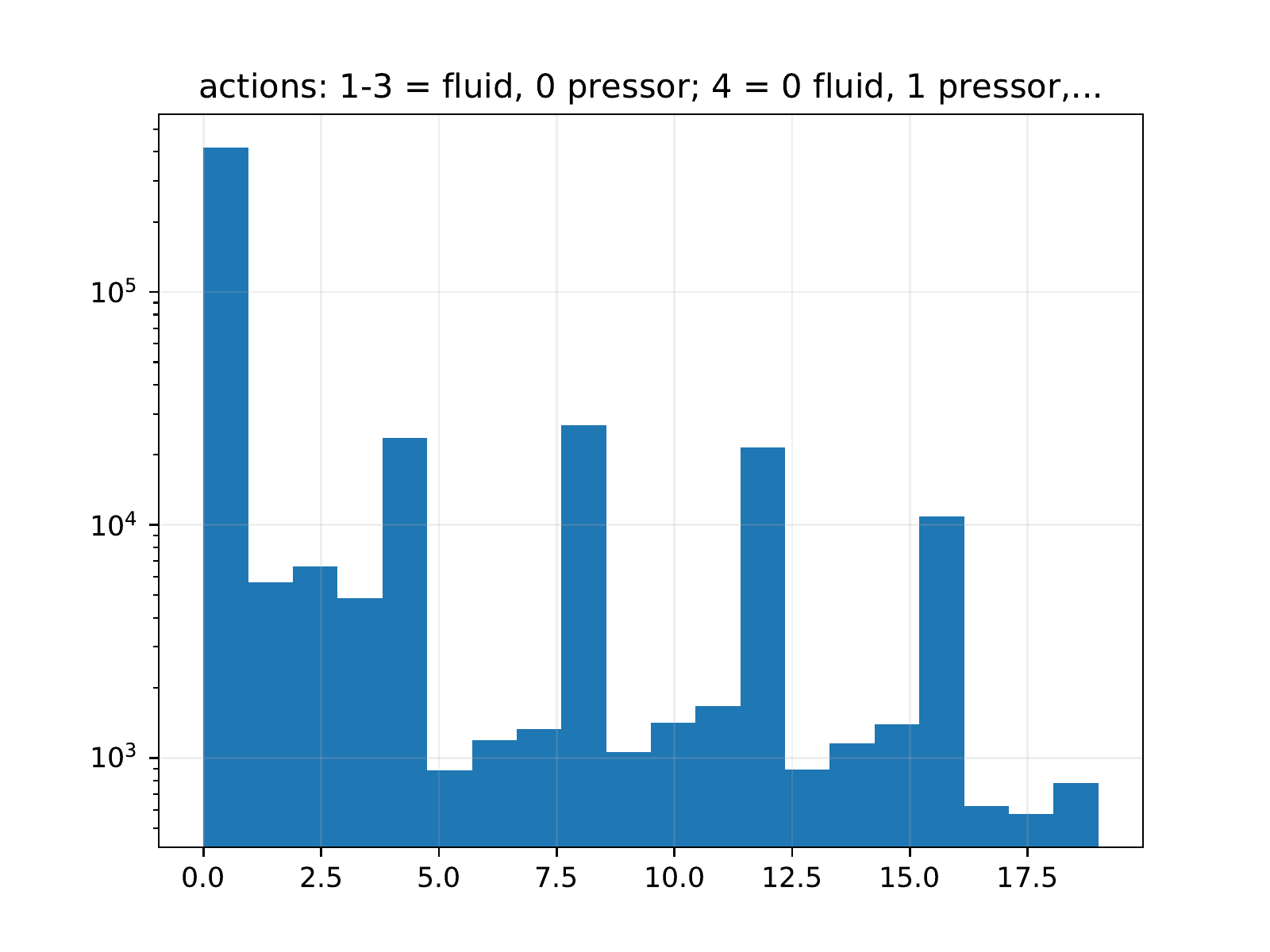}
}
\caption{Overall frequency of each action type in our dataset.}
\label{fig:data-action-amts}
\end{figure}

Lastly, Figure~\ref{fig:data-action-amts} shows the frequency by time point for how often each of the 20 different types of actions was administered. Roughly 85\% of all time points had no treatment administered.

\subsection{Reward Function Construction}
\label{app:mimic:rewards}

\begin{figure}[h!]
\centering{
\includegraphics[width=\textwidth]{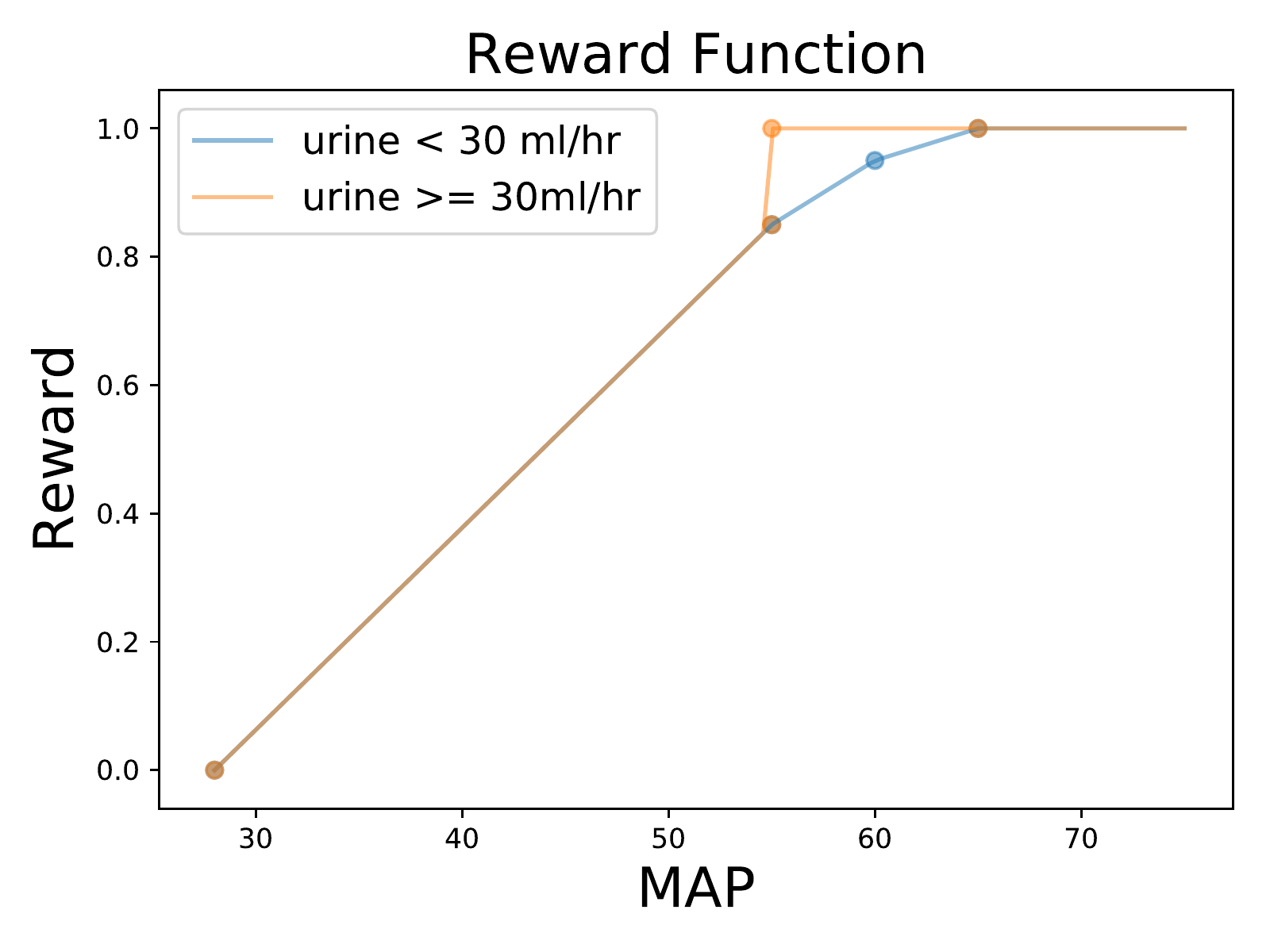}
}
\caption{The true reward function used in hypotension experiments. The algorithm is rewarded for keeping the the Mean Arterial blood Pressure (MAP) 65mmHg, a common target value in critical care.}
\label{fig:hypo-reward}
\end{figure}

We show reward function plots for the two reward functions used in this paper. Figure \ref{fig:hypo-reward} shows the MAP-based reward used in most of the work. Note the inflection points at 55 and 60 mmHg values. Also, patients who had adequate urine output had a lower threshold for MAP values that start to yield worse rewards, as clinically a modest urine output means the clinician is less worried about the precise MAP value unless it becomes very low.

\begin{figure}[h!]
\centering{
\includegraphics[width=\textwidth]{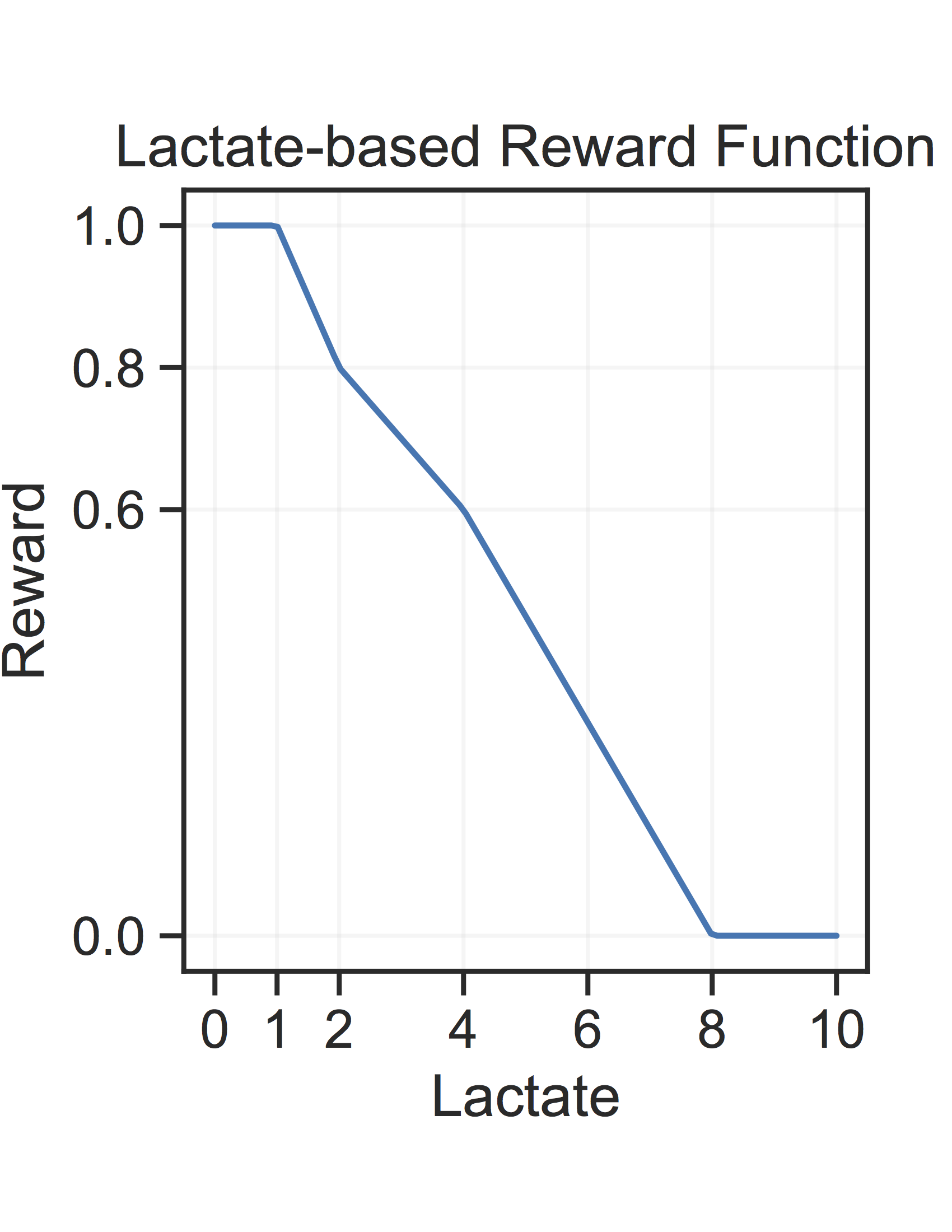}
}
\caption{Reward function based on lactate used for the reward re-specification experiment in the main text.}
\label{fig:new-lactate-reward}
\end{figure}

Figure \ref{fig:new-lactate-reward} shows the reward used for the reward re-specification experiment in the main paper, when we tested to what extent a model learned to yield good policies with respect to the MAP reward might generalize to this new reward. Clinically, higher lactate values indicate possible organ damage and are a sign of worsening illness.

\subsection{Learning the Behavior Policy}
\label{app:mimic:behavior}

We use the approach of \citet{raghu2018behaviour} to learn our behavior policy, and use a k-nearest-neighbors approach. Their work found that the calibration of the behavior policy is crucial for accurate OPE and that more complex models such as neural networks were often poorly calibrated. In consultation with our intensivist collaborator, we hand-constructed a distance function based on our observed variables, and used this to do kNN. 

We used a simple weighted Euclidean distance between observations, with a weight of 3 on creatinine, 2 on FiO$_2$, 3 on heart rate, 4 on lactate, 1 on platelets, 5 on urine output, 2.5 on total bilirubin, 5 on MAP, and 5 on GCS. Although not actually used as features in our models, we also considered 4 additional binary features that indicate the discrete vasopressor amount (if any) given at the last time point, and 3 binary features that indicate the discrete fluid amount administered at the last time point. All of these extra features received a weight of 5. We lastly added features that added the total raw amount of vasopressor and fluid given thus far in a trajectory, as well as in the past 8 hours; these 4 features also had a weight of 5. Concretely, we used $d(o, o') = \sum w_i (o_i - o'_i)^2$, with the weights $w_i$ listed previously and $i$ indexing observation variables. 

For performing kNN, we learned a behavior policy based on an observation's 100 nearest neighbors using our hand-crafted distance metric, and simply count up the actions performed by those neighbors to use as our estimate of behavior action probabilities. In rare cases where none of the 100 nearest neighbors correctly predicted the true next action taken, we reset the behavior policy to assign 3\% probability to the actual action that was taken.

We learn a different estimate of the behavior policy for each of the 5 folds of cross validation, using the same splits that were used by each of the later methods considered.

\subsection{Additional Qualitative Results}
\label{app:mimic:results}

\begin{figure*}[!t]
\centering
\setlength{\tabcolsep}{0.01cm}
\begin{tabular}{c c c}
  \includegraphics[height=0.19\textwidth]{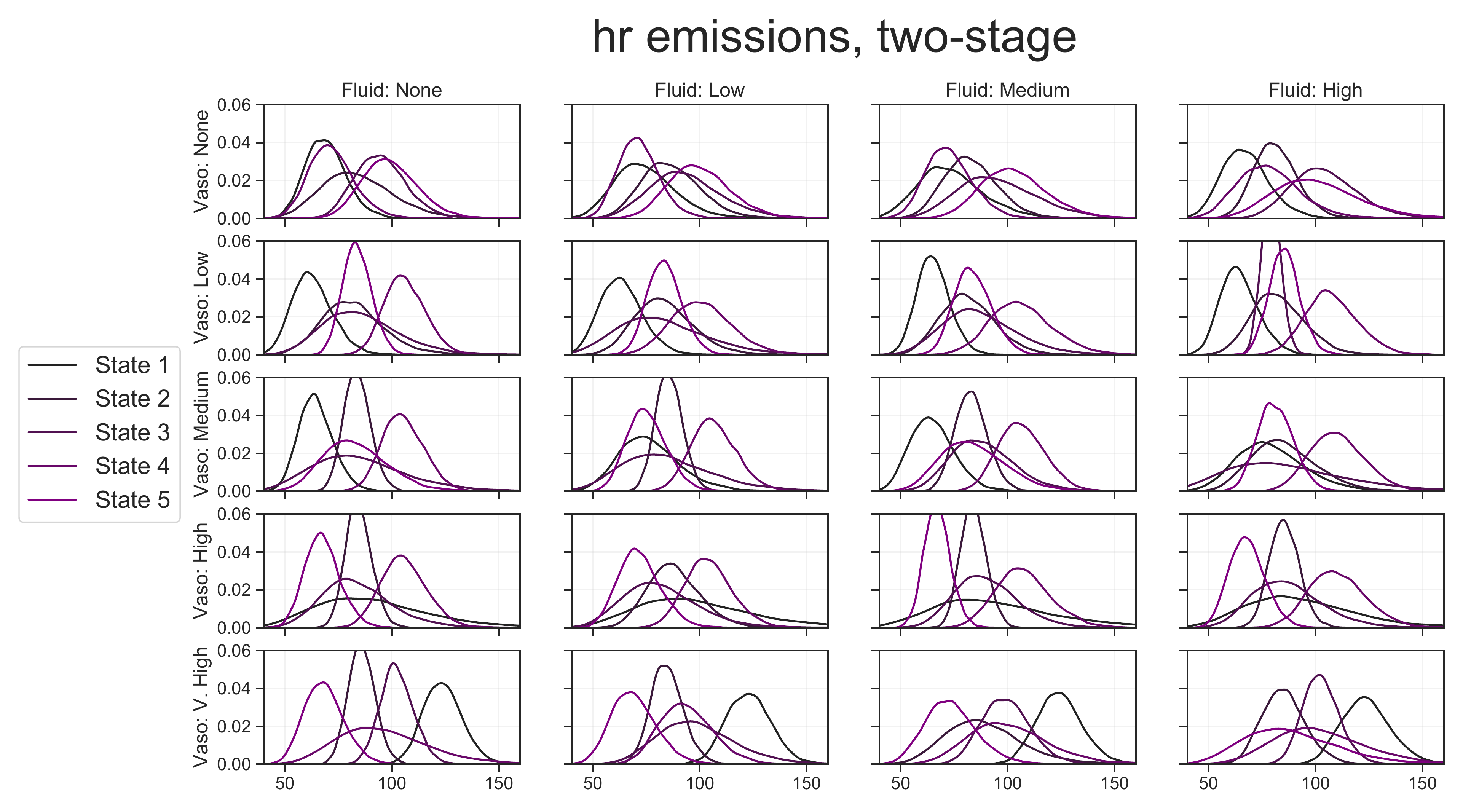}
&
  \includegraphics[height=0.19\textwidth]{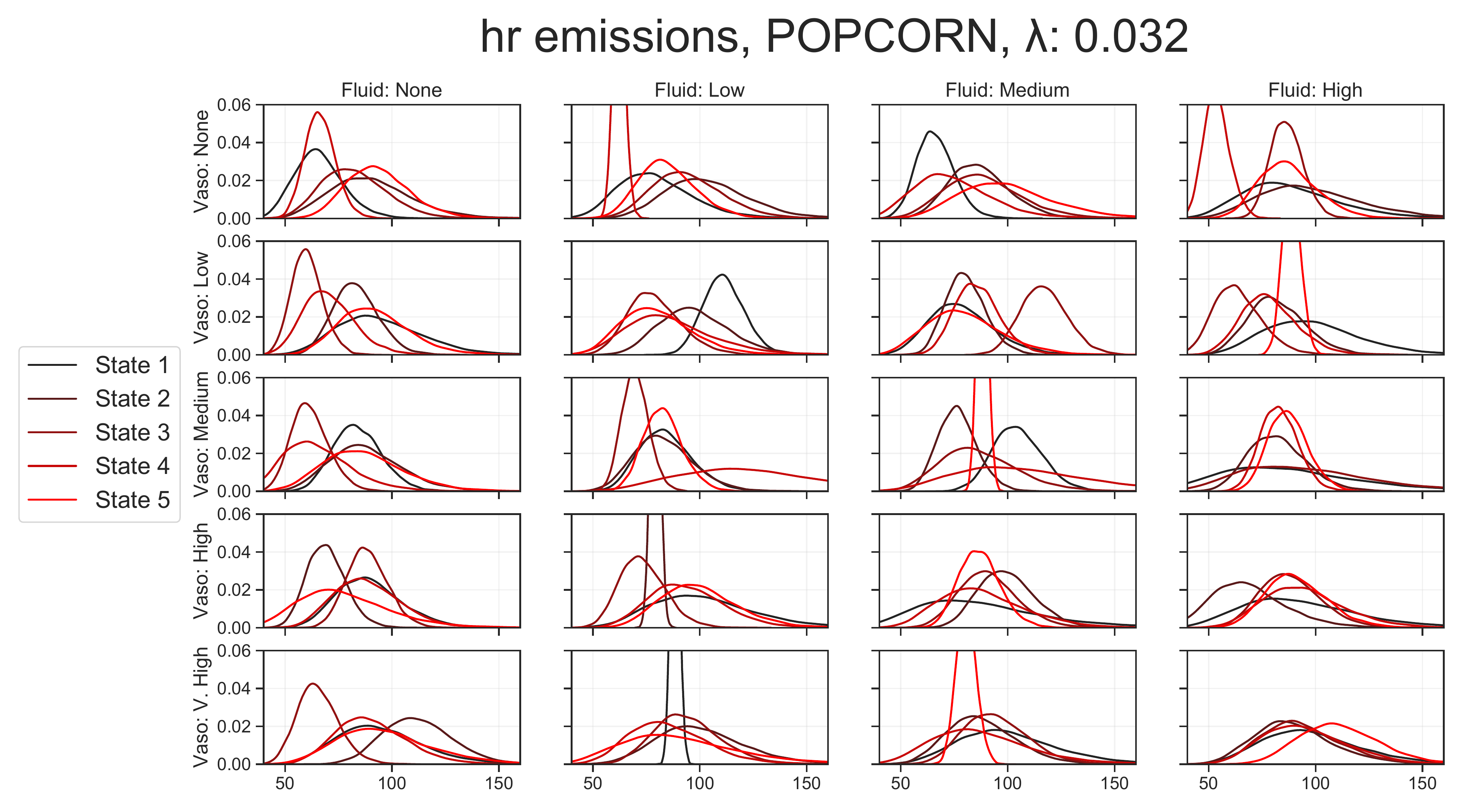}
&
  \includegraphics[height=0.19\textwidth]{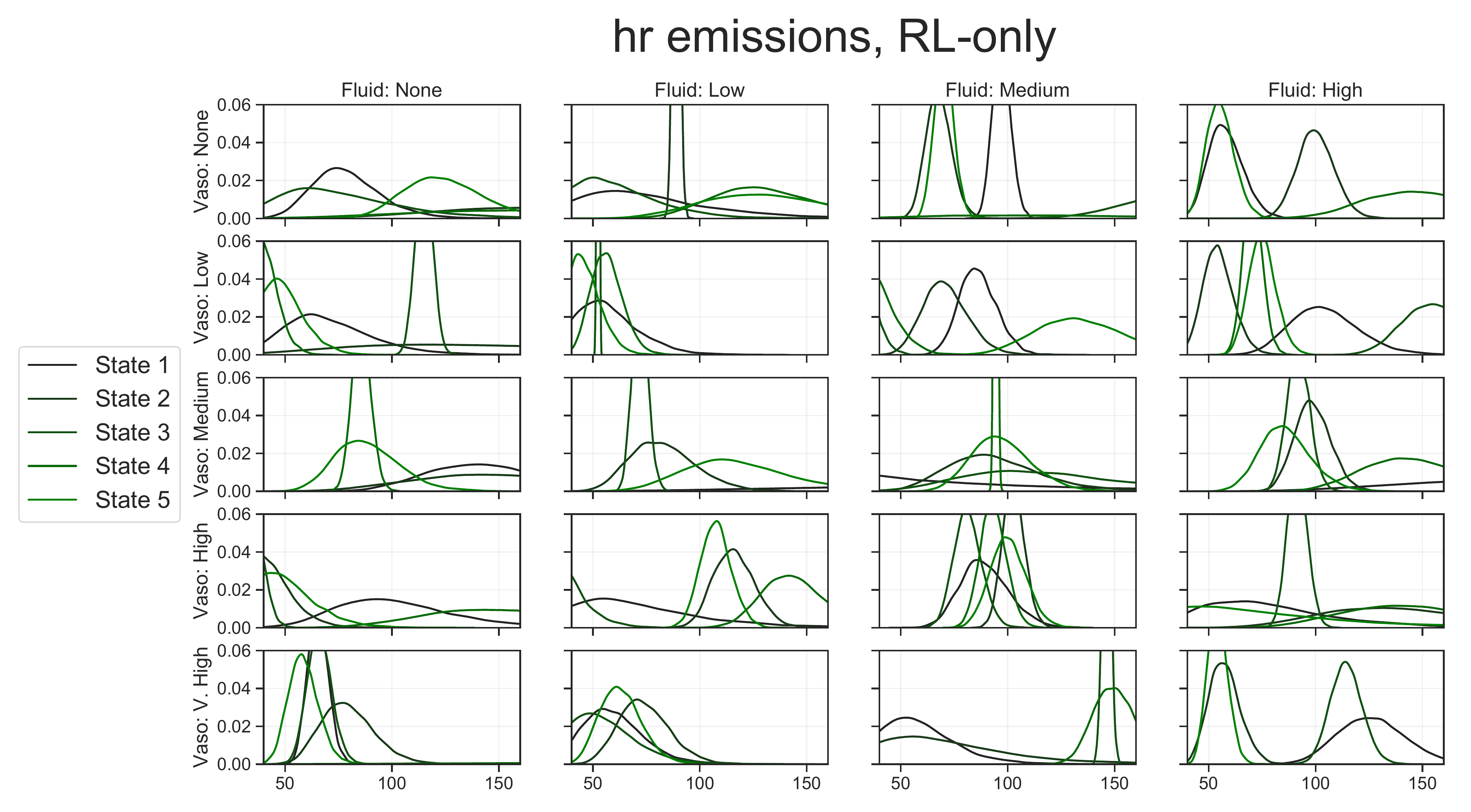}
\end{tabular}
\caption{
Visualization of learned heart rate distributions. 
\emph{Left:} 2-stage EM.
\emph{Middle:} POPCORN, $\lambda=0.032$.
\emph{Right:} Value-only.
Each subplot visualizes all $100$ learned distributions of heart rate values for a given method across the $20$ actions and $K=5$ states. Each pane in a subplot corresponds to a different action, and shows distributions across the $5$ states. Vasopressor actions vary across rows, and IV fluid actions vary across columns.
}
\label{fig:hr-dists}
\end{figure*}

\begin{figure*}[!t]
\centering
\setlength{\tabcolsep}{0.01cm}
\begin{tabular}{c c c}
  \includegraphics[height=0.19\textwidth]{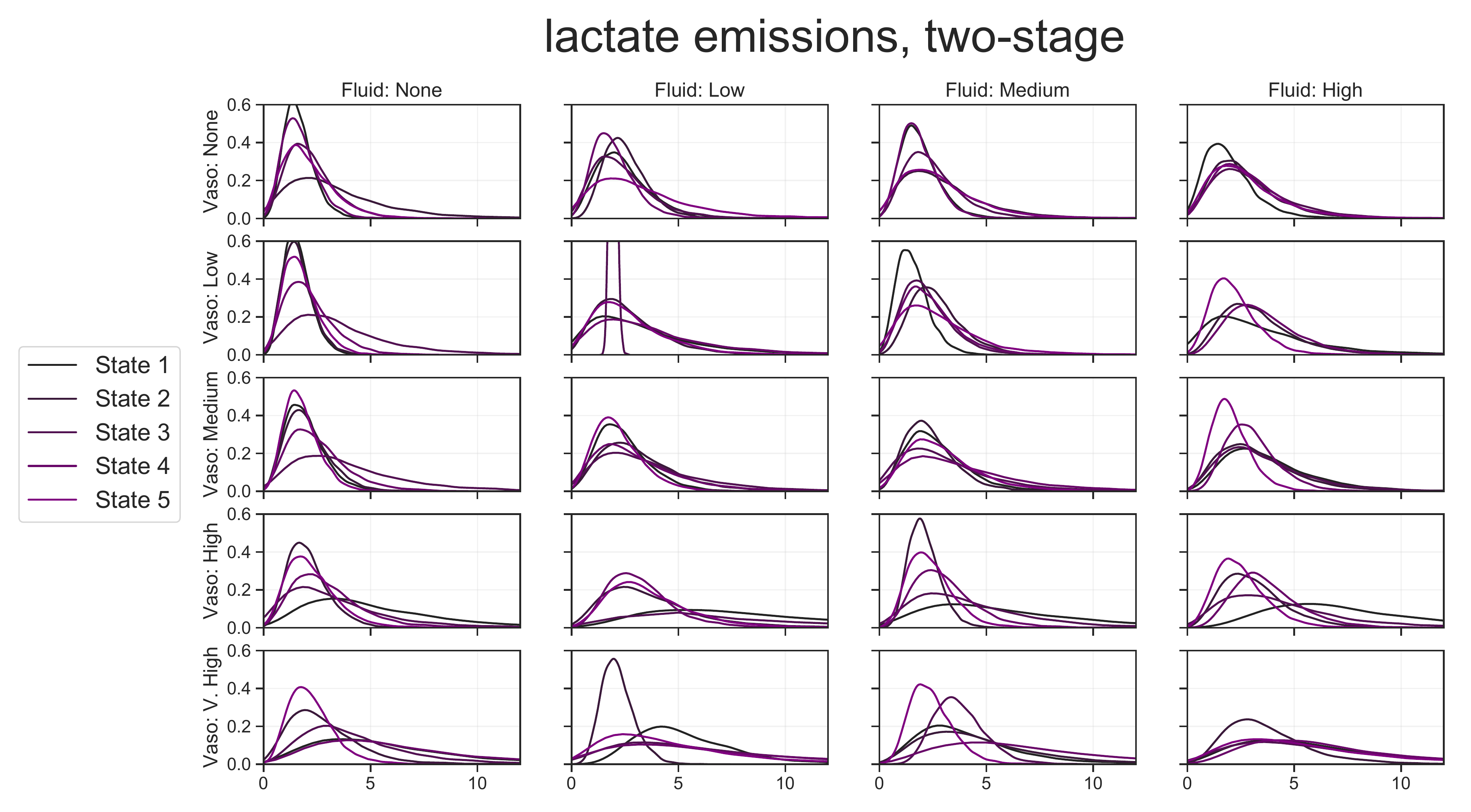}
&
  \includegraphics[height=0.19\textwidth]{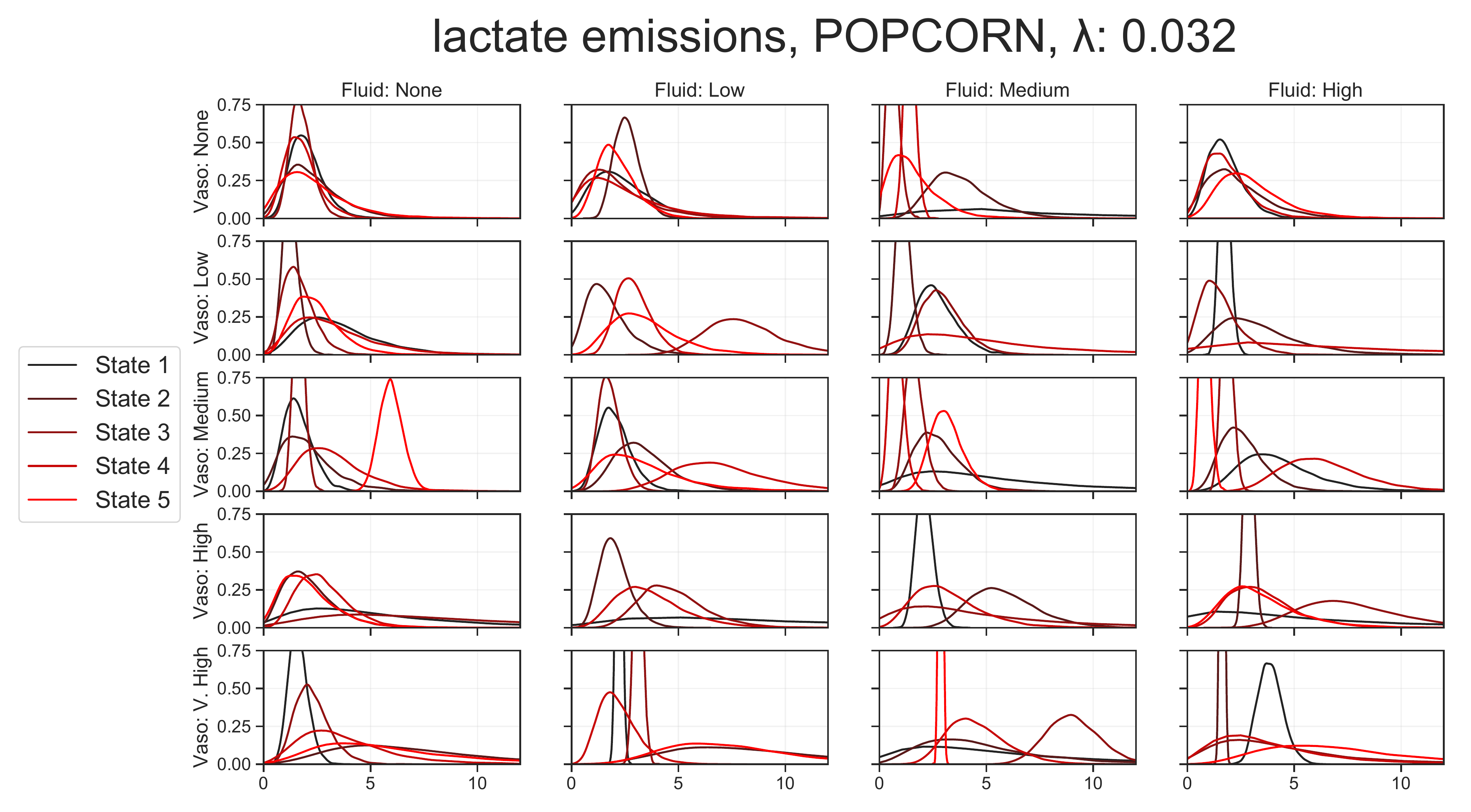}
&
  \includegraphics[height=0.19\textwidth]{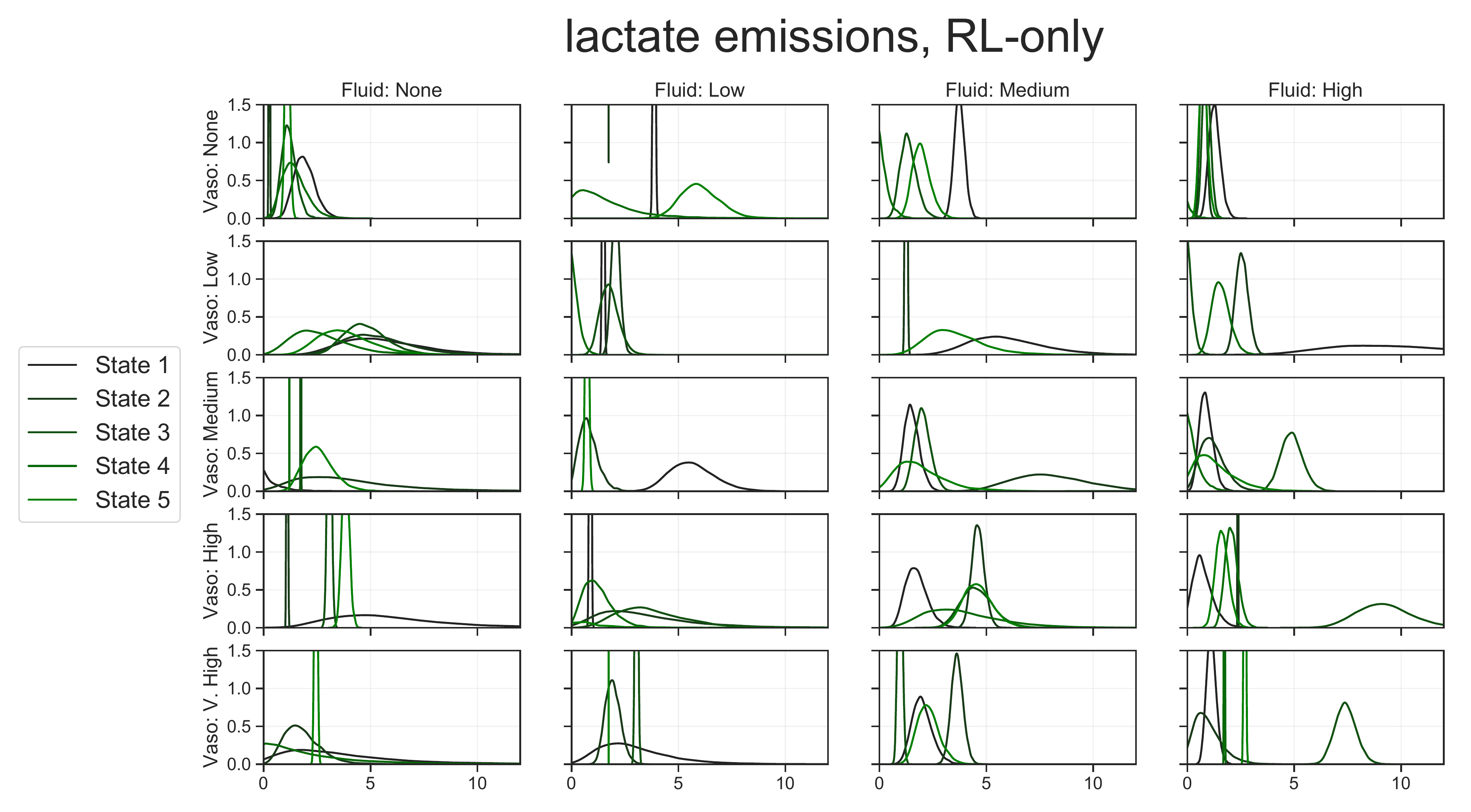}
\end{tabular}
\caption{
Visualization of learned lactate distributions. 
\emph{Left:} 2-stage EM.
\emph{Middle:} POPCORN, $\lambda=0.032$.
\emph{Right:} Value-only.
Each subplot visualizes all $100$ learned distributions of lactate values for a given method across the $20$ actions and $K=5$ states. Each pane in a subplot corresponds to a different action, and shows distributions across the $5$ states. Vasopressor actions vary across rows, and IV fluid actions vary across columns.
}
\label{fig:lactate-dists}
\end{figure*}

\begin{figure*}[!t]
\centering
\setlength{\tabcolsep}{0.01cm}
\begin{tabular}{c c c}
  \includegraphics[height=0.19\textwidth]{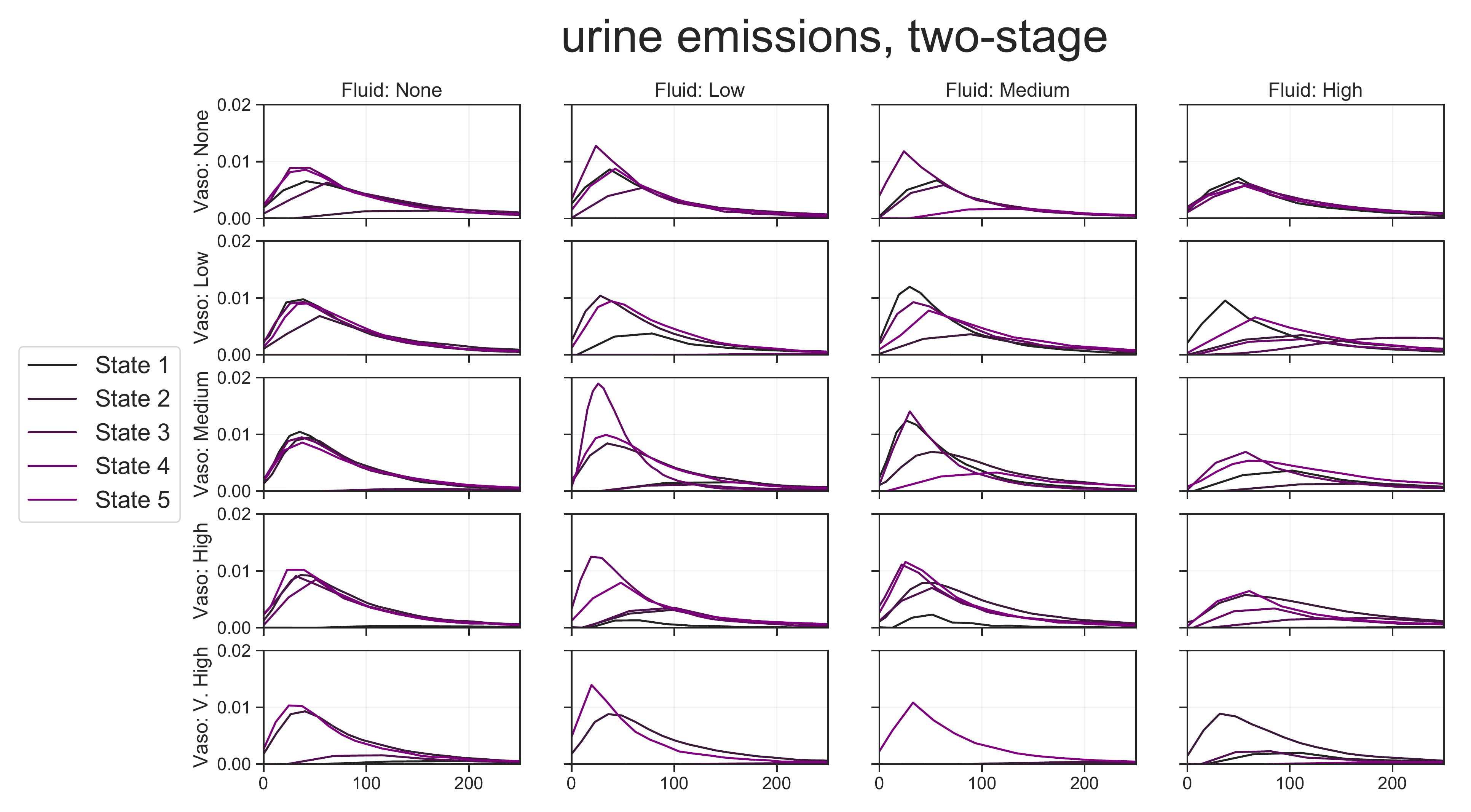}
&
  \includegraphics[height=0.19\textwidth]{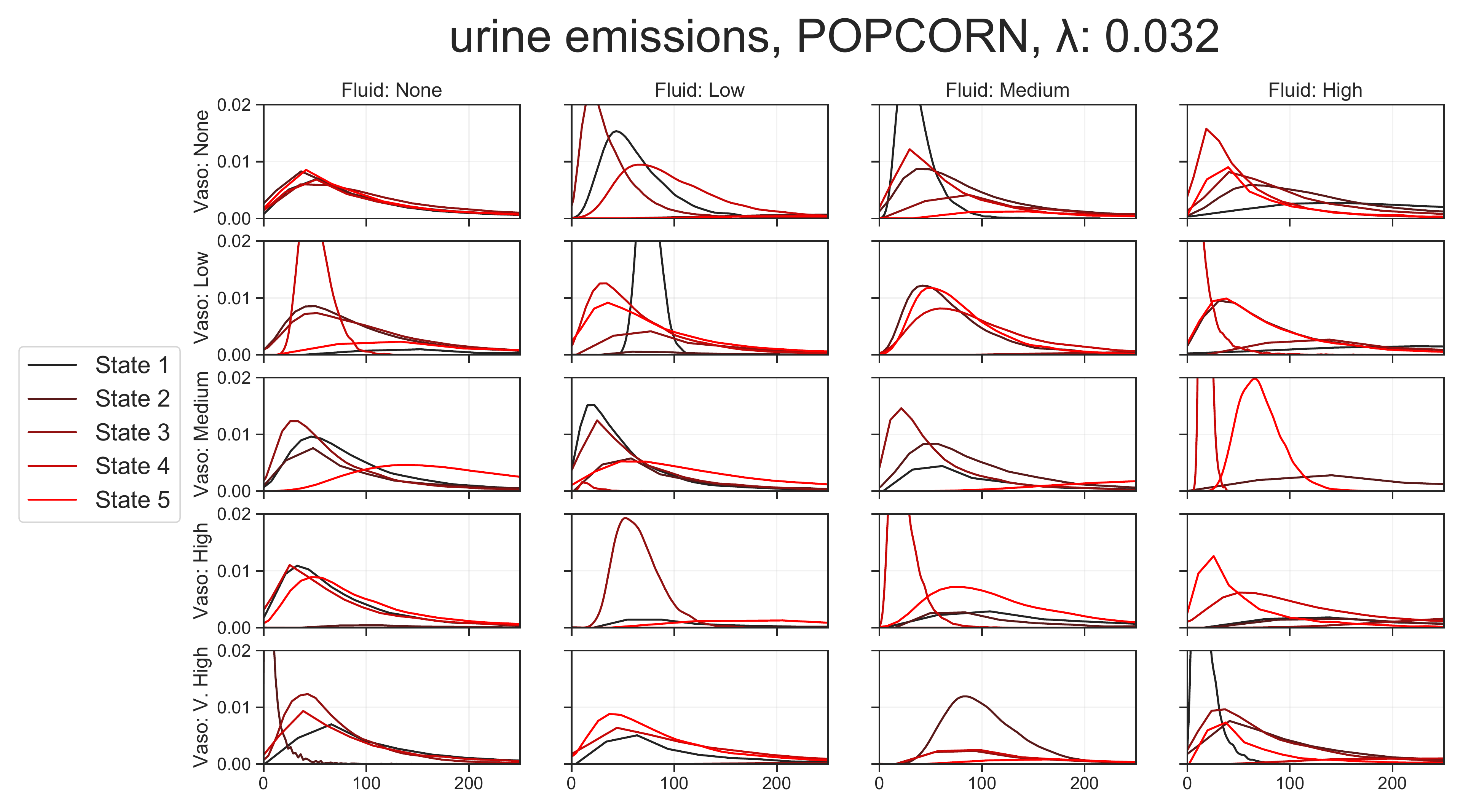}
&
  \includegraphics[height=0.19\textwidth]{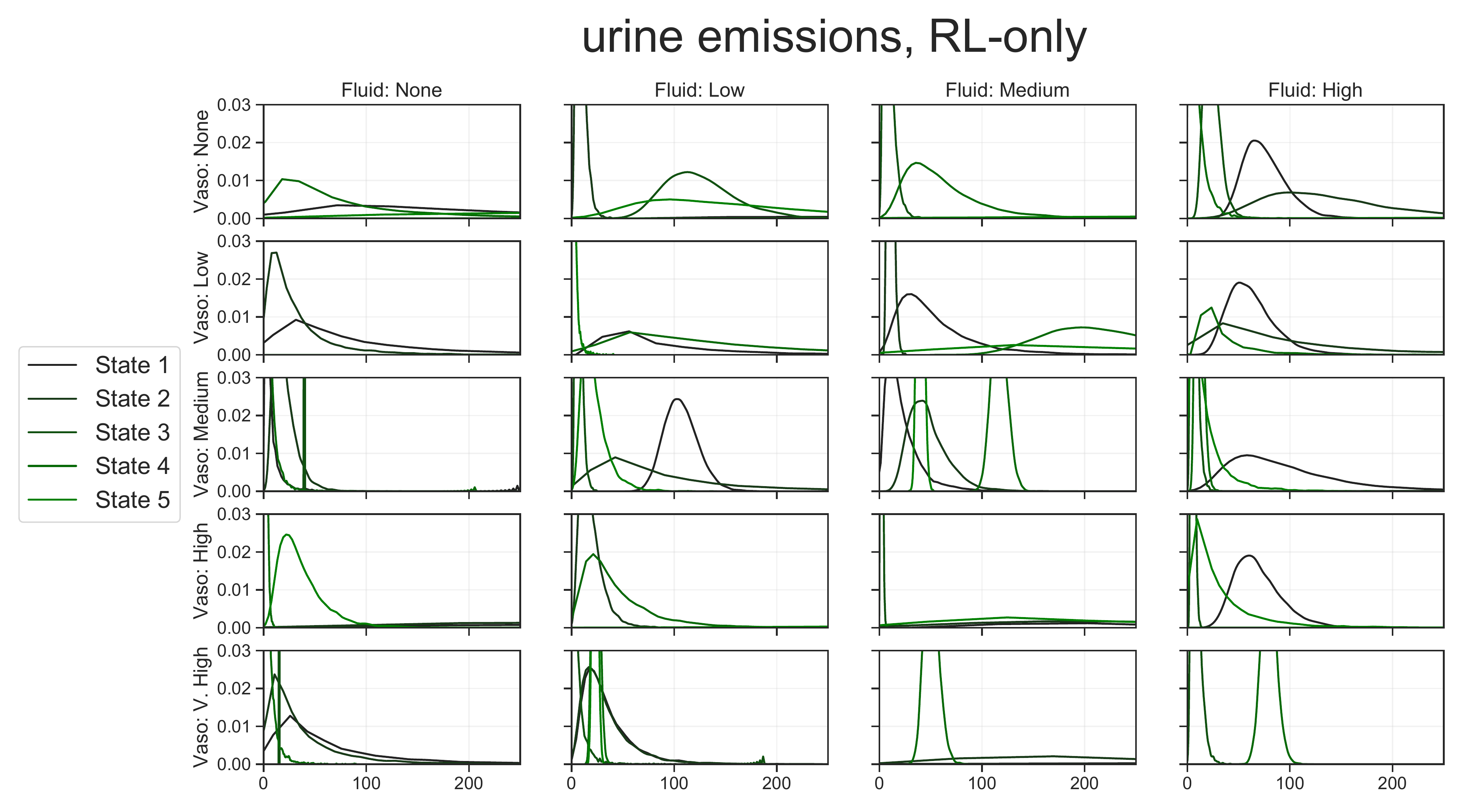}
\end{tabular}
\caption{
Visualization of learned urine output distributions. 
\emph{Left:} 2-stage EM.
\emph{Middle:} POPCORN, $\lambda=0.032$.
\emph{Right:} Value-only.
Each subplot visualizes all $100$ learned distributions of urine output values for a given method across the $20$ actions and $K=5$ states. Each pane in a subplot corresponds to a different action, and shows distributions across the $5$ states. Vasopressor actions vary across rows, and IV fluid actions vary across columns.
}
\label{fig:uo-dists}
\end{figure*}

Figures~\ref{fig:hr-dists},~\ref{fig:lactate-dists}, and~\ref{fig:uo-dists} show additional qualitative results about the learned models for POPCORN, 2-stage EM, and value-only, for the heart rate, lactate, and urine output variables. As in Figure~\ref{fig:MAP-dists} in the main text, we again see that the 2-stage approach largely learns states that exhibit very high overlap. Likewise, the value-only baseline learns states that are much more spread apart, and even occasionally learn bizarre distributions that are close to a point mass at one value. As expected, POPCORN learns models in between these two extremes, with diverse enough states to learn a good policy while also fitting the data decently well.

\end{document}